\documentclass{article}
\usepackage{times}
\usepackage{tikz}
\usepackage{multirow}
\usetikzlibrary{fit,positioning}
\usepackage{amsfonts}
\usepackage[toc]{appendix}
\usepackage{xr}
\usepackage{graphicx} 
\usepackage{subfigure} 
\usepackage{amsfonts,mathrsfs}
\usepackage{natbib}
\usepackage{algorithm}
\usepackage{algorithmic}

\usepackage{KTHleijonMath}
\newcommand{\bexp}{\mathfrak{exp}}
\newcommand{\blog}{\mathfrak{log}}
\newcommand{\sym}{\msf{sym}}
\newcommand{\yh}{\widehat{\vc Y}}
\renewcommand{\log}{\mr{log}}

\usepackage[accepted]{icml2019}

\icmltitlerunning{Constructing the Matrix Multilayer Perceptron and its Application to the VAE}

\begin{document} 
\twocolumn[
\icmltitle{Constructing the Matrix Multilayer Perceptron and its Application to the VAE}



\icmlsetsymbol{equal}{*}

\begin{icmlauthorlist}
\icmlauthor{Jalil Taghia}{to}
\icmlauthor{Maria B{\r{a}}nkestad}{to,to2}
\icmlauthor{Fredrik Lindsten}{to}
\icmlauthor{Thomas B. Sch{\"o}n}{to}
\end{icmlauthorlist}

\icmlaffiliation{to}{Department of Information Technology, Division of Systems and Control, Uppsala University, Uppsala, Sweden.}
\icmlaffiliation{to2}{SICS Computer Systems Laboratory, RISE Research Institutes of Sweden AB}

\icmlcorrespondingauthor{J. Taghia}{jalil.taghia@it.uu.se}
\icmlcorrespondingauthor{T. B. Sch{\"o}n}{thomas.schon@it.uu.se}

\icmlkeywords{}

\vskip 0.3in
]



\printAffiliationsAndNotice{}  
\begin{abstract} 
Like most learning algorithms, the multilayer perceptrons (MLP) is designed to learn a vector of parameters from data. However, in certain scenarios we are interested in learning structured parameters (predictions) in the form of symmetric positive definite matrices. Here, we introduce a variant of the MLP, referred to as the matrix MLP, that is specialized at learning symmetric positive definite matrices. We also present an application of the model within the context of the variational autoencoder (VAE). Our formulation of the VAE extends the vanilla formulation to the cases where the recognition and the generative networks can be from the parametric family of distributions with dense covariance matrices. Two specific examples are discussed in more detail: the dense covariance Gaussian and its generalization, the power exponential distribution. Our new developments are illustrated using both synthetic and real data.
\end{abstract} 
\section{Introduction}
\label{sec:intro}
We consider the problem of learning a symmetric positive definite (SPD) matrix in a nonlinear regression setting, ${ {\vc Y} = f(\vc X)}$. The input ${\vc X}$ is a matrix of arbitrary size which can be represented as a column vector without loss of generality. The output ${\vc Y}$, on the contrary, is a structured matrix in the form of an SPD matrix. Let the training set ${\mc D_{\mr{train}}=\{\vc Y_i, \vc X_i\}_{i=1}^n}$ include $n$ instances of such inputs and outputs. The task is to learn the function $f$ such that given an unseen test input ${{\vc X_{\star}}\in \mc D_{\mr{test}}}$, it produces the prediction of the corresponding output ${\widehat {\vc Y}}_{\star}$ under the constraint that $\widehat {\vc Y}_{\star}$ has to be an SPD matrix. 

Consider solving the problem using a multilayer perceptron (MLP) neural network \citep[e.g.,][]{Goodfellow2016}. The standard MLP  cannot straightforwardly be used, since as in most other neural network architectures, an MLP is designed to learn a vector of parameters from data without the consideration of any constraints. 
The objective is to design a nonlinear architecture which can learn the target outputs ${\widehat {\vc Y}}$ while satisfying the SPD  constraint across all layers. 
Our \textit{main contribution} is to show how to alter the architecture of the MLP in such a way that it not only respects the constraints, but also makes explicit use of them. 
We will achieve this by: 
1) Explicitly taking the non-Euclidean geometry of the underlying SPD manifolds \citep[e.g.,][]{Pennec2005} into account by designing a new loss function, and 2)  by deriving a new backpropagation algorithm \citep{Rumelhart1986} that respects the SPD nature of the matrices.
This new model will be referred to as the \emph{matrix multilayer perceptron (mMLP)}.
The mMLP makes use of positive-definite kernels to satisfy the SPD requirement across all layers. Hence, it provides a natural way of enabling deep SPD matrix learning.

We take a step-by-step approach in the development of the model. We first develop a simplified version of the resulting model that is designed for learning SPD matrices. We then extend this model into its most general form and show how it can be applied in connection to the VAE \citep{Kingma2014,Rezende2014}. More specifically, we replace the MLP in the vanilla VAE with the mMLP. This will crucially allow us to consider more general parametric families of distributions, in particular, those with dense covariance (dispersion) matrices.
Two concrete examples are considered: the \emph{dense covariance} multivariate Gaussian distribution and the multivariate power exponential (mPE) distribution \citep[e.g.,][]{Gomez1998}. 
Based on the parametric choice of the distributions we examine the effect of increasing model flexibility not only on the VAE's recognition network but also its generative network. This is achieved by relaxing the diagonality assumption on the covariance matrices and the Gaussian assumption. 
\section{Related Work}
\label{sec:related}
\paragraph*{SPD manifold metric.}
Earlier approaches for analyzing SPD matrices relied on the Euclidean space. Several recent studies suggest that non-Euclidean geometries such as the Riemannian structure may be better suited \citep[e.g.,][]{Arsigny2006,Pennec2005}. In this work, we consider the von Neumann divergence \citep[e.g.,][]{Nielsen2000} as our choice of the SPD manifold metric which is related to the Riemannian geometry. Previously, \citet{Tsuda2005} used this divergence in derivation of the matrix exponentiated gradients. Their work suggests its effectiveness for measuring dissimilarities between positive definite (PD) matrices. 
\vspace{-2ex}
\paragraph*{SPD manifold learning.}
There are multiple approaches towards the SPD matrix learning, via flattening SPD manifolds through tangent space approximations \citep[e.g.,][]{Tuzel2008,Fathy16}, 
mapping them into reproducing kernel Hilbert spaces \citep{Harandi2012b,Minh2014}, 
or geometry-aware SPD matrix learning \citep{Harandi2014}. 
While these methods typically follow shallow learning, the more recent line of research aims to design a deep architecture to nonlinearly learn target SPD matrices \citep{Ionescu2015,Huang2017ARN,Masci2015GeodesicCN,Huang2018}. Our method falls in this category but differs in the problem formulation. While the previous methods address the problem where the input is an SPD matrix and the output is a vector, we consider the reverse problem where the input is a matrix with an arbitrary size and the output is an SPD matrix. 
\vspace{-2ex}
\paragraph*{Backpropagation.} Our extension of the matrix backpropagation differs from the one introduced by \citet{Ionescu2015}. In their work, the necessary partial derivatives are computed using a two-step procedure consisting of first computing the functional that describes the variations of the upper layer variables with respect to the variations of the lower layer variables, and then computing the partial derivatives with respect to the lower layer variables using properties of the matrix inner product. In contrast, we make use of the concept of $\alpha$-derivatives \citep{Magnus2010} and its favorable generalization properties to derive a routine which \emph{closely} mimics the standard backpropagation. 
\vspace{-2ex}
\paragraph*{Flexible variational posterior in the VAE.}
An active line of research in the VAE is related to designing flexible variational posterior distributions that preserve dependencies between latent variables.
In this regard, the early work of \cite{Rezende2014} proposed the use of the rank-1 covariance matrix with a diagonal correction. Although computationally attractive, this makes for a poor approximation of the desired dense covariance matrix. Another approach is to induce dependencies between latent variables by considering a dependence on some auxiliary variables \citep{Maale2016} or assuming hierarchical structures \citep{Ranganath2016,Tran2016}. Finally, an alternative approach towards achieving flexible variational posteriors is based on the idea of normalizing flow \citep{Rezende2015,Kingma2016}. In this work, we take a different approach toward increasing the posterior flexibility. This is achieved by learning dense covariance matrices via the mMLP model and relaxing the Gaussian assumption via the mPE distribution. While the approach taken by \citet{Kingma2016} toward learning dense covariance matrices solves an overparameterized problem, in our formulation, we make explicit use of kernels, within the mMLP architecture, to learn the dense covariance matrices.  
\vspace{-3ex}
\section{Preliminaries}
\label{sec:preli}
\paragraph{Matrix $\alpha$-derivative.}
Throughout this work we adopt the \emph{narrow} definition of the matrix derivatives known as the $\alpha$-derivative \citep{Magnus2010} in favor of the broad definition, $\omega$-derivative. The reason for this is that the $\alpha$-derivative has better generalization properties. This choice turned out to be crucial in the derivation of the mMLP's backpropagation routine which involves derivatives of matrix functions w.r.t. the matrix of variables. The $\alpha$-derivative and some of its properties are introduced in Appendix~\ref{app:alpha}.
\vspace*{-2ex}
\paragraph{Bregman matrix divergences.}
Let us restrict ourselves to the domain of SPD matrices. Furthermore, let ${\mc F(\vc X):\mbb R\p{d\times d}\rightarrow \mbb R}$ be a real-valued strictly convex differentiable function of the parameter domain and ${\mathscr {F}(\vc X)=\vc \nabla_{\vc X} \mc F(\vc X)}$, where $\vc \nabla_{\vc X} \mc F(\vc X)$ denotes the gradient w.r.t. the matrix. Then the \emph{Bregman divergence} between $\vc X$ and $\widetilde{\vc X} $ is defined as \citep[e.g.,][]{Kulis2009}
\begin{align}
\label{eq:bergman}
\!{\Delta_{\mc F} (\widetilde{\vc X} || \vc X)\! :=\! \mc F (\widetilde{\vc X}) \!-\! \mc F({\vc X}) - \msf {tr} ((\widetilde{\vc X}\! - \!\vc X) \mathscr F({\vc X})^\top)}.
\end{align}
Bregman divergences are non-negative, definite, and in general asymmetric. There are several choices for the function $\mc F$ \citep[e.g.,][]{Sra2016}. The most common choice is probably ${\mc {F}(\vc X)=-\log \mr{det}}(\vc X)$, which leads to the Stein divergence \citep{Stein1956}, or commonly known as the LogDet divergence (refer to Appendix~\ref{app:Stein} for details). However, in this work, we argue in the favor of the \emph{von Neumann entropy}, also known as the \emph{quantum relative entropy (QRE)} \citep[e.g.,][]{Nielsen2000} as the choice of function. A numerical example is discussed in Section~\ref{sec:example1} which highlights the advantage of the von Neumann divergence over the Stein divergence as the choice of the SPD manifold metric within the mMLP architecture.

Using the von Neumann entropy as our choice of function in \eqref{eq:bergman}, we arrive at:
${\mc F (\vc X) = \msf{tr}(\vc X\blog \vc X - \vc X)}$, where $\blog $ denotes the matrix logarithm---for an SPD matrix $\vc A$, it is computed using ${\blog\vc A= \vc V \mr{diag}(\log \bc\lambda)\vc V^\top}$, where $\vc V $ and $\bc\lambda$ are the matrix of eigenvectors and the vector of eigenvalues from the eigendecomposition of $\vc A$. The Bregman divergence corresponding to this choice of function is known as the von Neumann divergence, ${\Delta_{\mc F} (\widetilde{\vc X} || \vc X) = \msf {tr}(\widetilde{\vc {X}} \blog \widetilde{\vc X} - \widetilde{\vc X} \blog{\vc X} - \widetilde{\vc X} + {\vc X})}$.
Throughout, we consider the cases where the parameters are normalized so that: ${\msf {tr}(\vc X) = \msf{tr}(\widetilde{\vc X})=1}$. The normalized von Neumann divergence is given by
\begin{align}
\label{eq:qre}
\Delta_{\mr {QRE}} (\widetilde{\vc X} || \vc X) = \msf {tr}(\widetilde{\vc {X}} \blog \widetilde{\vc X} - \widetilde{\vc {X}} \blog{\vc {X}}).
\end{align}
 
\vspace*{-4ex}
\section{Matrix Multilayer Perceptron}
\label{sec:mmlp}
We first construct the basic form of the mMLP suitable for learning SPD matrices. Next, we
 construct its general form which will be applied in connection to the VAE.
\vspace*{-1ex}
\subsection{The Basic Form of the mMLP}
\label{sec:mmlp1}
\vspace*{-1ex}
\paragraph{Activation matrix function.}
Let ${\vc Z=(\vc z_1, \ldots, \vc z_{d})}$ denote a matrix of variables ${\vc z_i\in \mbb R\p {d}}$. The activation function $\mc K(\vc Z)$ defines a matrix function in the form of ${[\mc K(\vc Z)]_{i,j} = \kappa(\vc z_{i}, \vc z_{j}) }$, ${\forall i, j \in\{1, \ldots, d\}}$,
where $\kappa$ is some differentiable activation function outputting scalar values. In the following, we restrict ourselves to the kernel functions which form PD activation matrix functions. Irrespective of the functional form of $\kappa$, we will---mostly for numerical reasons and partly for the fact that our loss function in \eqref{eq:loss_qre} will make use of the normalized von Neumann divergence---need to normalize the resulting kernel matrix. 
This can be achieved by enforcing the trace-one constraint,
\begin{align}
{\mc H(\vc Z) = {\mc K(\vc Z)}/{\msf{tr}(\mc K(\vc Z))}},
\end{align}
 where $\mc H$ denotes a differentiable PD activation matrix function of trace one.
 Without loss of generality, throughout this work, we use the Mercer sigmoid kernel \citep{Carrington2014} defined as
 \begin{align}
 \label{eq:msk}
{\kappa({\vc z_i, \vc z_j) = {\mr{tanh} (\alpha \vc z_i + \beta)} \odot \mr{tanh} (\alpha \vc z_j + \beta)}  },
 \end{align} 
where $\alpha$ and $\beta$ denote the slope and the intercept, respectively. Furthermore, $\odot$ denotes the dot product. In all experiments, we use default values of ${\alpha=1}$ and ${\beta=0}$. 

\vspace{-2ex}
\paragraph{Model construction.}
Let ${\vc X\in \mbb{R}^{p_1\times p_2}}$ indicate the input matrix and ${\vc Y\in \mbb{R}^{d_0\times d_{0}}}$ indicate the corresponding output matrix, an SPD matrix of trace one. The mMLP of $j$ hidden layers is shown as $\msf{mMLP\!:\!\vc X\! \rightarrow\! \widehat{\vc Y}}$ and constructed as
\begin{align}
\label{eq:mlpa}
\begin{split}
&\begin{cases}
\widehat{\vc Y} = \mc H(\vc Z_0),\\
\vc Z_0 = \vc W_0 \vc H_1 \vc W_0^\top + \vc B_0,
\end{cases}
 \\
&\begin{cases}
\vc H_l = \mc H(\vc Z_l),  \\
\vc Z_l = \vc W_l \vc H_{l+1} \vc W_l^\top+ \vc B_l,
\end{cases}
\\
&
\begin{cases}
\vc H_{j+1} = \mc H(\vc Z_{j+1}),  \\
\vc Z_{j+1} = \vc W_{j+1} \mr{vec}\vc X (\vc W_{j+1}\vc 1_{p_1p_2})^\top  + \vc B_{j+1},
\end{cases}
\end{split}
\end{align}
where the pair of ${\vc W_l\!\in\! \mbb R^{d_l\times d_{l+1}}, \forall {0\leq\! l\!\leq j},}$ and ${\vc W_{j+1} \!\in\! \mbb R^{d_{j+1}\times p_1p_2}}$ are the weight matrices, ${\vc B_l\!\in\! \mbb R^{d_l\times d_l}, \forall {0\!\leq \!l\leq j\!+\!1},}$ are the bias matrices, ${\vc Z_l\in \mbb R^{d_l\times d_l}, \forall {0\leq \!l\!\leq j\!+\!1}}$, are the latent input matrices, and ${\vc H_l\in \mbb R^{d_l\times d_l}, \forall {1\leq l\leq j+1},}$ are latent output SPD matrices of trace one.

In the construction of \eqref{eq:mlpa}, we have ensured that $\vc H_l$ are SPD matrices of trace one \emph{across all layers} as opposed to only at the output layer. The idea is to propagate the nonlinearities introduced via the SPD activation matrix functions through all  layers. This design choice turned out to be more effective than the alternative, and arguably simpler, design where the SPD requirement is met only at the output layer. We will discuss this further in Section~\ref{sec:shvsde}, where we also present an illustrative numerical example.

\vspace{-2ex}
\paragraph{Loss function.} 
We consider the normalized von Neumann divergence \eqref{eq:qre} as the base for the loss function. The von Neumann divergence is asymmetric. However, it can be symmetrized by using the fact that the von Neumann entropy of trace one follows the class of generalized quadratic distances \citep{Nielsen2007}. Hence, we define the loss function as
\begin{multline}
\label{eq:loss_qre}
\!\!\!\ell_{\mr{QRE}}(\widehat{\vc Y}, \vc Y) 
=  \frac{1}{2}(\Delta_{\mr {QRE}} (\yh || \vc Y) + \Delta_{\mr {QRE}} (\vc Y ||\yh)),
\end{multline}
where $\Delta_{\mr {QRE}}$ is given by \eqref{eq:qre}. The $\alpha$-derivative of $\ell_{\mr{QRE}}$ involves taking partial derivatives through the eigendecomposition. In Appendix~\ref{sec:der_loss}, we derive a method for analytically computing the derivative of $\ell_{\mr{QRE}}$. 

\vspace{-2ex}
\paragraph*{Optimization.} The remaining steps are feed-forward computation, backpropagation, and learning, as in the standard MLP. However, here, the backpropagation requires taking derivatives with respect to the matrix functions. These steps are described in Appendix~\ref{app:basic_mlp}.
\vspace{-1ex}
\subsection{The General Form of the mMLP}
\label{sec:mmlp2}
We now discuss a general version of the mMLP which produces both a vector and an SPD matrix as outputs. One possible application of this model is for heteroscedastic multivariate regression (we do not pursue this application here). Another application is within the VAE formulation, as we discuss in more detail in Section~\ref{sec:vae}.
\vspace{-2ex}
\paragraph{Model construction.}
Let ${\vc X\in \mbb{R}^{p_1\times p_2}}$ denote the input matrix. The corresponding outputs in this case are: ${\vc Y\in \mbb{R}^{d_0\times d_{0}}}$ which is an SPD matrix of trace one, and ${\vc y\in \mbb{R}^{r_0}}$. 
The mMLP of $j$ hidden layers is shown as $\msf{mMLP\!:\!\vc X\! \rightarrow\! \{ \widehat{\vc y}, \widehat{\vc Y}\}}$ and constructed as:
\begin{align}
\label{eq:mmlpb}
\begin{split}
&\begin{cases}
\widehat{\vc y} = \mathfrak{h} (\vc z_0),\\
\vc z_0 = \vc C_0\yh \vc A_0 \vc h_1 + \vc b_0,\\
\widehat{\vc Y} = \mc H(\vc Z_0), \\
\vc Z_0 = \vc W_0 \vc H_1 \vc W_0^\top + \vc B_0,
\end{cases}
 \\
&\begin{cases}
\vc h_l = \mathfrak{h} (\vc z_l),\\
\vc z_l = \vc C_l\vc H_{l}\vc A_l \vc h_{l+1} + \vc b_l,  \\
\vc H_l = \mc H(\vc Z_l), \\
\vc Z_l = \vc W_l \vc H_{l+1} \vc W_l^\top+ \vc B_l,
\end{cases}
\\
&
\begin{cases}
\vc h_{j+1} = \mathfrak{h} (\vc z_{j+1}),\\
\vc z_{j+1} = \vc C_{j+1}\vc H_{j+1} \vc A_{j+1} \bc 1 + \vc b_{j+1}, \\
\vc H_{j+1} = \mc H(\vc Z_{j+1}), \\
\vc Z_{j+1} = \vc W_{j+1} \mr{vec}\vc X (\vc W_{j+1}\vc 1_{p_1p_2})^\top  + \vc B_{j+1},
\end{cases}
\end{split}
\end{align}
where 
${{\vc h_{l} \!\in\!{\mbb R^{r_{l}}}}, {\vc H_l\!\in\! \mbb R^{d_l\times d_l}}, \forall {1\!\leq \!l\leq \!j\!\!+\!\!1\!}}$, ${\vc z_l, \vc b_l \!\in \!{\mbb R^{r_l}}}$, ${\vc Z_l,\! \bc B_l \! \in\! \mbb R^{d_l\times d_l}}$, ${\vc C_l\!\in\!{\mbb R^{r_{l} \times d_l }}},\! \forall  {0\!\leq l \!\leq\! j\!+\!1}$, ${\vc A_l\!\in\!{\mbb R^{d_l\times r_{l+1}}}\!}$, $ {\vc W_l \!\in \!\mbb R^{d_l\times d_{l+1}}}, \ \forall {0\leq \!l\!\leq j}$. 
 Just as in the standard MLP, $\mathfrak{h}$ is an activation function of choice, e.g., the hyperbolic tangent function. 

\vspace{-2ex}
\paragraph*{Loss function.}
The loss function ${\ell(\widehat{\vc Y}, \widehat{\vc y}, \vc Y, \vc y)} $ needs to be designed with the specific application in mind. 
In the context of the VAEs, the default loss function is provided by the lower bound on the marginal likelihood. We will explore this choice in Section~\ref{sec:vae}.
\vspace{-2ex}
\paragraph*{Optimization.}
The remaining steps of feed-forward computation, backpropagation, and learning, are all described in Appendix~\ref{app:gen_mlp}. 

\vspace*{-2ex}
\section{Exploiting the mMLP within the VAE}
\label{sec:vae}

\subsection{Background and Problem Formulation}
\label{sec:back_vae}
Let ${\{\vc x\h i\}_{i=1}^n}$ indicate the set of i.i.d. observations on the real space ${\vc x\h i\!\in\!\mbb R^{d}}$. It is assumed that the data are generated by some random process involving a continuous latent variable ${\vc s\in \mbb R^{k}}$ admitting a joint distribution ${p_{\theta, \pi} (\vc x, \vc s)= p_{\theta}(\vc x \given \vc s) p_{\pi}(\vc s)}$ parametrized by $\pi$ and $\theta$. Here, ${p_{\theta}(\vc x \given \vc s)}$ is the generative model which is also known as the decoder, and ${p_{\pi}(\vc s)}$ is the prior. 
Let ${q_{\phi}(\vc s\given \vc x)}$ indicate the recognition model also known as the encoder, parametrized by $\phi$. The distribution ${q_{\phi}(\vc s\given \vc x)}$ approximates the intractable true posterior $p_{\theta, \pi}(\vc s\given \vc x)$. \citet{Kingma2014} introduced a method based on variational inference \citep[refer to][for a recent review] {Blei2017} for learning the recognition model parameters $\phi$ jointly with the generative model parameters $\theta$ and $\pi$. 
This is done by maximizing a lower bound $\mc L (\theta, \pi, \phi)$ on the marginal log-likelihood of the data, also known as the evidence lower bound (ELBO),
\begin{multline}
\label{eq:ll}
\mc L = \mbb E_{q_{\phi}(\vc s\mid \vc x)}[\log~{p_{\theta}(\vc x \given \vc s)}] - \Delta_{\mr{KL}}(q_{\phi}(\vc s\given \vc x) || p_{\pi}(\vc s)), 
\end{multline}
where $\Delta_{\mr{KL}}(q||p)$ is the Kullback-Leibler divergence (KLD) between $q$ and $p$. 
\vspace{-2ex}
\paragraph*{The vanilla VAE.} The parametric form of the recognition model is assumed to be a diagonal Gaussian distribution, ${q_{\phi}(\vc s\mid \vc x) = \mc N(\bc \mu_q, \mr{diag}(\bc\sigma_q^2))
}$ with parameters $\bc \mu_q$ and $\bc\sigma_q$ which are outputs of an MLP as a function of $\vc x$.
 Similarly for the observations on the continuous real space, the parametric form of the generative model is assumed to take on a diagonal Gaussian distribution, ${p_{\theta}(\vc x\given \vc s) = \mc N(\bc \mu_p, \mr{diag}(\bc\sigma_p^2))
}$ with parameters $\bc \mu_p$ and $\bc\sigma_p$ which are outputs of an MLP as a function of $\vc s$.
\vspace{-2ex}
\paragraph*{The diagonality assumption.} The VAE formulation is general and as pointed out by \citet{Kingma2014}, the reason they insisted on using a diagonal covariance matrix on the Gaussian is to simplify the resulting inference. However, this choice comes with the cost of losing the dependencies between latent variables. It is fair to say that most extensions of the vanilla VAE make the same choice. The key reason for this choice is the computational complexity associated with maintaining a dense matrix.

Irrespective of the computational complexity, 
the question we are interested in is that whether there would be any significant gain in relaxing this assumption? 
When it comes to the recognition model, we know for a fact that it is never possible to have overfitting. Thus, it can only be advantageous to increase the expressiveness of the  recognition network by expressing the posterior via more flexible family of parametric distributions, e.g., dense covariance Gaussian distributions. There is in fact an active line of research that is focused on recapturing some of the lost dependencies between the latent variables (see Section~\ref{sec:related} for references).
On the other hand, when it comes to the generative network, there might be the possibility of overfitting: The argument is that if the generative network is flexible enough, the VAE model may simply choose to ignore the latent variables, forcing the posterior to approach the prior. However, the question is how realistic this scenario really is? Perhaps more importantly, should this be a reason to instead force the generative network to take on a simple parametric form, e.g., the diagonal Gaussian distribution? 

To get insights into these questions, we will consider several model alternatives with various degrees of flexibility on both the recognition model and the generative model. For this purpose, we will in the subsequent section replace the MLP with the mMLP which allows us to construct models with the dense covariance matrices. Two parametric families of distributions are considered: the multivariate Gaussian distribution and the mPE distribution. The latter drops the Gaussian assumption and provides a simple way to construct models with the various degree of flexibilities. 

\subsection{VAE via mMLP}
\label{sec:vae_mmlp}
We first introduce the parametric distributions we have chosen to work with (Section~\ref{sec:ParFamDist}) and then we proceed with the model constructions (Section~\ref{sec:vae_model_const}), and finally we derive the estimators (Section~\ref{sec:Estimators}). 
\vspace*{-1ex}
\subsubsection{Parametric Families of Distributions}
\label{sec:ParFamDist}
\paragraph*{Trace-one Gaussian distribution.}
The output layer of the mMLP model in \eqref{eq:mmlpb} operates under the trace-one constraint. It is of course possible to drop the trace one constraint from the output layer, but we would then also be forced to use a different kernel function 
in that layer. Instead, we find it easier to work with a reparameterized Gaussian distribution with a trace-one covariance matrix. This allows us to use the same choice of kernel function~\eqref{eq:msk} across all layers.

 For a random variable ${\bc \vartheta\in \mbb R^{d}}$, the trace-one Gaussian distribution is shown as ${\mc N_{\mr{tr1}}\!(\bc \vartheta; \bc \mu, \!\bc \Omega, \!\eta)}$ where ${\bc \mu\in \mbb R^{d}}$ is the mean, ${\eta\in \mbb R^+}$ is the scale, and ${\bc \Omega\in\mbb{R}^{d\times d}}$ is the trace-one covariance matrix, ${\mr {tr}(\bc \Omega)=1}$.
  Refer to Appendix~\ref{App:t1gauss} for the exact functional form of the distribution and its stochastic representation. 
\vspace*{-2ex}
\paragraph{Trace-one power exponential distribution.}
\label{sec: MGGD} 
A generalized variant of the multivariate Gaussian distribution is provided by the mPE distribution \citep[e.g.,][]{Gomez1998}. As in the case of the Gaussian distribution, we find it easier to work with a reparameterized representation where the dispersion matrix is of trace one. Imposing this trace-one constraint, for a random variable ${\bc \vartheta\in\mbb R^{d}}$, the trace-one mPE distribution can be expressed as
\begin{align}
\vspace*{-2ex}
&\!\mc{E}_{\mr{tr1}}(\bc \vartheta; \bc\mu, \vc \Omega, \eta, \alpha, \beta)\!\!=\!\!\frac{c({\alpha, \beta})}{(\mathrm{det}(\eta\bc\Omega))^{\frac{1}{2}}}\! \exp\!\left\{\!\! -\!\frac{1}{2}\!\!\left( \!\frac{t(\bc \vartheta;\bc \mu, \!\vc \Omega)}{\alpha\eta} \!\right)^{\!\beta}\!\right\}, \notag \\
& c({\alpha, \beta}) = \frac{\beta \Gamma(\frac{d}{2})} {\pi^{\frac{d}{2}} \Gamma(\frac{d}{2\beta})2\p{\frac{d}{2\beta}} \alpha\p{\frac{d}{2}}},
\\
& t(\bc\vartheta;\bc \mu, \vc \Omega):=(\bc \vartheta-\bc \mu)^\top \vc \Omega^{-1} (\bc \vartheta-\bc \mu), \quad \mr{tr}(\bc \Omega)=1. \notag
\end{align}
The pair of ${\alpha\in\mbb R^{+}}$ and ${\beta\in\mbb R^{+}}$ are the scale and shape parameters of the density, $\bc \mu$ is the mean vector, and ${\vc \Omega}$ is a ${d\times d}$ symmetric real dispersion matrix where $\mr{tr}(\bc \Omega)=1$. The parameter $\eta$ has the same role as in the trace-one Gaussian distribution. 

Figure~\ref{fig:pdf} shows the probability density function of the distribution for various values of $\alpha$ and $\beta$ (for the case of ${d=2}$). As a special case, the mPE includes the Gaussian distribution: For ${\alpha=1}$ and ${\beta=1}$, the trace-one mPE distribution corresponds to the trace-one multivariate Gaussian distribution. 
 
\emph{\textbf{Stochastic representation (reparametrization trick).}}
Let ${\bc\vartheta\in\mbb R^{d}}$, ${\bc\vartheta\sim {\mc {E}_{\mr{tr1}}}(\bc\mu, \vc\Omega, \eta, \alpha, \beta)}$, and ${\phi=\{\bc\mu, \vc\Omega, \eta, \alpha, \beta\}}$. The density admits a known stochastic representation
\begin{align}
\label{eq:rt_mpe}
\vspace*{-2ex}
\! \! \!\bc\vartheta \stackrel{\mathsf d}{=} \mc T_{\mc E_{\mr{tr1}}}(\bc\nu, \varsigma, \bc\vartheta ; \phi)= \bc \mu + \varsigma \vc \Phi \bc\nu, \  \alpha\eta\vc \Omega = \vc \Phi \vc \Phi^\top,
\end{align}
where $\stackrel{\mathsf d}{=}$ denotes the equality in distribution, $\bc\nu$ is a random vector uniformly distributed on the unit-sphere of $\mbb R^{d}$ such that ${\bc\nu^\top \bc\nu=1}$, and $\varsigma$ is a positive scalar random variable distributed according to a known gamma distribution,
\begin{align}
\label{eq:gamma_app}
\vspace{-2ex}
{\varsigma^{2\beta}\sim \mc G({d}/{2\beta}, 2)}.
\end{align} The shape of the gamma unfortunately depends on $\beta$. As there is no known stochastic representation of the gamma distribution, it ultimately causes difficulties when we need to take the derivative of the random sample. There are techniques which can be used for this purpose \citep{Ruiz2016,Naesseth2017}. However, in our case, $\beta $ is a ``second-level'' parameter in the sense that it appears in \eqref{eq:gamma_app} and not directly in \eqref{eq:rt_mpe}. For our specific case, we found that approximating the gamma with a normal distribution as the limiting distribution of the gamma to be sufficiently accurate. More specifically, 
\begin{align}
\label{eq:gauss_app}
{\varsigma^{2\beta}\sim \mc N({d}/{\beta}, {2d}/{\beta})} \ \Rightarrow \ \varsigma^{2\beta} \stackrel{\mathsf d}{=} {d}/{\beta} + \epsilon\sqrt{{2d}/{\beta}}, \ 
\end{align}
where ${\epsilon\sim\mc N(0, 1)}$. The error in the approximation \eqref{eq:gauss_app} is expected to decrease as the shape parameter of the gamma distribution increases. A pessimistic upper bound is ${\left({d}/{2\beta}\right)^{-{1}/{2}}}$ which follows from the Berry-Ess\`{e}en theorem. In our numerical evaluations, we found that the use of \eqref{eq:gauss_app} instead of \eqref{eq:gamma_app} in \eqref{eq:rt_mpe} is quite well suited, surprisingly even in extreme cases where ${d=2}$ and $\beta$ is large (see the numerical comparisons in Appendix~\ref{app:gamma_vs_gauss}).


\emph{\textbf{Practical considerations.}}
The pair of $\alpha$ and $\beta$ control the tail and the shape of the distribution (see Figure~\ref{fig:pdf}). Very large and very small values of $\alpha$ and $\beta$ might be undesirable---for one, they pose numerical challenges. In practice, these parameters can be restricted within a range by choosing suitable output activation functions. In the experiments in Section~\ref{sec:exp_vae}, we choose to bound them conservatively as: ${{0.5}\leq \alpha, \beta \leq 1.5}$, using the sigmoid function. 

\subsubsection{Constructing Two Models}
\label{sec:vae_model_const}
\paragraph*{Gaussian model.}
The parametric form of the recognition model is assumed to take on a trace-one Gaussian distribution, ${q_{\phi}(\vc s \given \vc x) = \mc N_{\mr{tr1}}(\bc \mu_q, \bc \Omega_q, \eta_q)}$ with parameters as the output of an mMLP as a function of $\vc x$, 
${\msf{mMLP}\!:\!\vc x\! \rightarrow\! \{ ({\bc \mu_q}, \log \eta_q), \bc\Omega_q\}}$.
 Similarly, the generative model is assumed to take on a trace-one Gaussian distribution in the form of ${p_{\theta}(\vc x\given \vc s) = \mc N_{\mr{tr1}}(\bc \mu_p, \bc \Omega_p, \eta_p)}$ with parameters as the output of an mMLP as a function of $\vc s$, ${\msf{mMLP}\!:\!\vc s\! \rightarrow\! \{ ({\bc \mu}_p, \log \eta_p), \bc\Omega_p\}}$. 
 \vspace*{-2ex}
\paragraph*{Power exponential model.} 
We can improve the flexibility of the recognition and the generative models by generalizing the Gaussian model to the mPE model. Here, it is assumed that the generative model takes on a trace-one mPE distribution in the form of ${p_{\theta}(\vc x\given \vc s) = \mc E_{\mr{tr1}}(\bc \mu_p, \bc \Omega_p, \eta_p, \alpha_p, \beta_p)}$ with parameters as the output of an mMLP as a function of $\vc s$, ${\msf{mMLP}\!:\!\vc s\! \rightarrow\! \{ ({\bc \mu_p}, \log \eta_p, \alpha_p, \beta_p), \bc\Omega_p\}}$. There are two alternatives when it comes to the recognition model: 
\begin{itemize}
\vspace{-2ex}
\item[1)] To assume the same parametric choice for the recognition model as the generative model and define: ${q_{\phi}(\vc s \given \vc x) \!=\! \mc E_{\mr{tr1}}(\bc \mu_q, \bc \Omega_q, \eta_q, \alpha_q, \beta_q)}$,~with parameters as the output of an mMLP as a function of $\vc x$, that is ${\msf{mMLP}\!:\!\vc x\! \rightarrow\! \{ ({\bc \mu_q}, \log \eta_q, \alpha_q, \beta_q), \bc\Omega_q\}}$. The disadvantage of this choice is that the KLD between the posterior and the prior will become analytically intractable. Thus, the estimator needs to rely on Monte Carlo sampling for the evaluation of the KLD term. This could increase the variance of the estimator. The advantage of this choice is that it gives additional flexibility to the recognition model which is desirable. 
\item[2)] \vspace{-1.ex} To restrict the recognition model to the dense Gaussian model and define: ${q_{\phi}(\vc s \given \vc x) = \mc N_{\mr{tr1}}(\bc \mu_q, \bc \Omega_q, \eta_q)}$ with parameters as the output of an mMLP as a function of $\vc x$, 
${\msf{mMLP}\!:\!\vc x\! \rightarrow\! \{ ({\bc \mu_q}, \log \eta_q), \bc\Omega_q\}}$. This choice leads to an analytically tractable KLD computation. 
\vspace{-2ex}
\end{itemize}
We will consider both cases in our evaluation. Based on the choice of the generative and the recognition networks, various models can be constructed. Table~\ref{tb:abb} summarizes the list of model variants considered in this work. For all these models, the prior is assumed to follow a standard Gaussian distribution as ${p_{\pi}(\vc s) = \mc N(\bc 0, \bc I)}$. 
\subsubsection{Estimators} 
\label{sec:Estimators}
Given ${\{\vc x\h i\}_{i=1}^n}$, an estimator of the lower bound of the full dataset can be constructed based on mini-batches of size $m$ as ${\widetilde{\mc L}_m\approx\frac{n}{m} \sum_{i=1}^m \widetilde{\mc L}(\theta, \pi, \phi;\vc x\h i)}$, where ${\widetilde{\mc L}(\theta, \pi, \phi;\vc x\h i)}$ is the estimate of the lower bound \eqref{eq:ll} using $r$ Monte Carlo samples. Based on the choice of the recognition model, the estimator is different. In the case of the Gaussian recognition model, $\widetilde{\mc L}$ is computed from
\begin{multline}
\label{eq:est_n}
\vspace{-3ex}
\widetilde{\mc L}(\theta, \pi, \phi;\vc x\h i) = \frac{1}{r}\sum_{l=1}^r\log~{p_{\theta}(\vc x\h i \given \vc s\h {i,l})} \\ - \Delta_{\mr{KL}}(q_{\phi}(\vc s\mid \vc x\h i) || p_{\pi}(\vc s)), 
\end{multline}
where ${\vc s\h {i,l} = \mc T_{\mc N_{\mr{tr1}}}(\bc \epsilon\h {i, l}, \vc x\h i ; \phi)}$ is given by \eqref{eq:rep}. In the case of the mPE recognition model, 
\begin{multline}
\label{eq:est_mpe}
\vspace{-3ex}
\!\!\!\!\!\widetilde{\mc L}(\theta,\! \pi, \!\phi;\vc x\h i) \!= \!\frac{1}{r}\sum_{l=1}^r\!\log~{p_{\theta}(\vc x\h i \given \vc s\h {i,l})}  \\ + \log p_{\pi}(\vc s\h {i,l}) - \log q_{\phi}(\vc s\h {i,l}\given \vc x\h i),\vspace{-3ex}
\end{multline}
where ${\vc s\h {i,l}\!\!= \!\!\mc T_{\mc E_{\mr{tr1}}}\!(\varsigma \h {i, l}, \!\bc \nu\h {i, l}, \!\vc x\h i; \phi )}$ is given by \eqref{eq:rt_mpe}.

Next, we need to take the $\alpha$-derivatives of the estimators \eqref{eq:est_n} and \eqref{eq:est_mpe} which are given in Appendix~\ref{app:estimator}.
\begin{table}[t!]
\vspace{-3ex}
\caption{Model specifications and abbreviations.}
\setlength{\tabcolsep}{3pt}
\renewcommand{\arraystretch}{1.5}
\tiny
\begin{tabular}{l*{3}{l}}
Model              & ${p_{\theta}(\vc x\given \vc s)}$ & ${q_{\phi}(\vc s \given \vc x)}$ &  Estimator \\
\hline
${\mc N_\mr{d}\mc N_\mr{d}}$ & $\mc N_{\mr{tr1}}(\bc \mu_p, \bc \Omega_p^{\mr{diag}}, \eta_p)$ &  $\mc N_{\mr{tr1}}(\bc \mu_q, \bc \Omega_q^{\mr{diag}}, \eta_q)$ & \eqref{eq:est_n} \\
${\mc N_\mr{d}\mc N_\mr{f}}$ & $\mc N_{\mr{tr1}}(\bc \mu_p, \bc \Omega_p^{\mr{diag}}, \eta_p)$ &  $\mc N_{\mr{tr1}}(\bc \mu_q, \bc \Omega_q^{\mr{full}}, \eta_q)$& \eqref{eq:est_n}  \\
${\mc N_\mr{f}\mc N_\mr{d}}$ & $\mc N_{\mr{tr1}}(\bc \mu_p, \bc \Omega_p^{\mr{full}}, \eta_p)$ &  $\mc N_{\mr{tr1}}(\bc \mu_q, \bc \Omega_q^{\mr{diag}}, \eta_q)$& \eqref{eq:est_n}  \\
${\mc N_\mr{f}\mc N_\mr{f}}$ & $\mc N_{\mr{tr1}}(\bc \mu_p, \bc \Omega_p^{\mr{full}}, \eta_p)$ &  $\mc N_{\mr{tr1}}(\bc \mu_q, \bc \Omega_q^{\mr{full}}, \eta_q)$& \eqref{eq:est_n} \\
${\mc E_\mr{f}\mc N_\mr{f}}$ & $\mc E_{\mr{tr1}}(\bc \mu_p, \bc \Omega_p^{\mr{full}}, \eta_p, \alpha_p, \beta_p)$ &  $\mc N_{\mr{tr1}}(\bc \mu_q, \bc \Omega_q^{\mr{full}}, \eta_q)$& \eqref{eq:est_n}\\
${\mc E_\mr{f}\mc E_\mr{f}}$ & $\mc E_{\mr{tr1}}(\bc \mu_p, \bc \Omega_p^{\mr{full}}, \eta_p, \alpha_p, \beta_p)$ & $\mc E_{\mr{tr1}}(\bc \mu_q, \bc \Omega_q^{\mr{full}}, \eta_q, \alpha_q, \beta_q)$ & \eqref{eq:est_mpe}
\end{tabular}
\vspace{-3ex}
\label{tb:abb}
\end{table}

 \section{Experiments}
 Our experiments are divided into two parts. The first part is on the empirical validation of the mMLP model in a supervised task of learning SPD matrices using synthetic data. The second part evaluates the VAE models in an unsupervised task using real data.
\subsection{SPD Matrix Learning}
\label{sec:spd_mtx}
\subsubsection{The Choice of Loss Function}
\label{sec:example1}
Consider the problem of learning SPD matrices on synthetic data using the mMLP model of \eqref{eq:mlpa}. The objectives are to validate the model, and to evaluate the effect of the choice of the loss function on the performance. In particular how the choice of the SPD manifold metric under Bergman matrix divergence affects the learning. 

For this purpose, here, we consider two candidate loss functions in the family of Bergman matrix divergences. The first candidate is the loss function based on the normalized von Neumann divergence ${\ell_{\mr {QRE}}(\widehat{\vc Y}, \vc Y)}$ given by \eqref{eq:loss_qre}. The second candidate is based on the symmetrized Stein divergence ${\ell_{\mr{Stein}}(\widehat{\vc Y}, \vc Y)}$ given by
Eq.~\eqref{eq:loss_stein}. These two candidates are related to the Riemannian geometry. 
For the sake of comparison, we also consider the quadratic loss ${\ell_{\mr{quad}}(\widehat{\vc Y}, \vc Y)=\mr{tr}((\yh - \vc Y)(\yh - \vc Y)^{\top}})$ which is related to the Euclidean geometry.
\vspace{-2ex}
\paragraph{Example~1.}
Consider the set ${\mc D_{\mr{train}}\!\!=\! \{\vc X_i,\! \vc Y_i\}_{i=1}^{n_{\mr{train}}}}$ of inputs ${\vc X_i\in \mbb R^{20\times 1}}$ and corresponding SPD matrix outputs ${\vc Y_i\in \mbb{R}^{d_0\times d_0}}$ which are in this case dense PD covariance matrices (refer to Appendix~\ref{app:example1} for details on the data generation). The goal is to estimate the covariance matrices ${\widehat{\vc Y}}$ associated to the input vectors from the unseen test set, ${{\vc X}_i \!\in \!\mc D_{\mr{test}}}$. The training size is varied between ${n_{\mr{train}}\!=\!\{20, 100\}}$ samples. The analysis is carried out for ${d_0\!=\!\{10, 20\}}$.
Two examples of the test outputs $\vc Y_i$ for ${d_0\!=\!20}$ are shown in Figure~\ref{fig:example1}-A. 

The mMLP models \eqref{eq:mlpa} are trained using 3 layers (20 units per layer) under three choices of loss functions: $\ell_{\mr{QRE}}$, $\ell_{\mr{Stein}}$, and $\ell_{\mr{quad}}$. The \emph{only} difference here is the loss function.
 Refer to Appendix~\ref{app:example1} for additional details on the mMLP initialization.
The performance is evaluated on the test set, ${n_{\mr{test}}=10^3}$, in terms of all three losses as the error measures, shown as $E_{\mr{QRE}}, E_{\mr{Stein}}, E_{\mr{quad}}$. 

Table~\ref{tb:ex1} summarizes the results of the evaluation. The first observation is that the quality of estimates differs considerably depending on the choice of the loss function.
The loss function ${\ell_{\mr {QRE}}}$ that takes into account the geometry of the SPD matrices outperforms the one based on the Euclidean geometry, ${\ell_{\mr {qaud}}}$. Between the two choices of Bergman divergences (${\ell_{\mr {QRE}}}$ and ${\ell_{\mr {Stein}}}$), the ${\ell_{\mr {QRE}}}$ is clearly the best performer: It performs consistently well in terms of the various error measures, and shows robustness even in cases where the training data are limited.
Figures~\ref{fig:example1}-B,C,D visualize the predicted covariance matrices for the case of ${d_0=20}$ and ${n_{\mr{train}}=20}$ for two test samples.

For the sake of comparison, we also solved the same problem using the standard MLP as a regressor with the quadratic loss. To meet the SPD requirement, we simply used the Cholesky decomposition. In general, the performance was quite poor in comparison to the mMLP model (refer to Appendix~\ref{app:ex1mlp} for additional details).
\begin{table}[t!]
 \vspace{-3ex}
  \tiny
  \caption{SPD matrix learning using the mMLP (Example~1).}
   \vspace{-3ex}
  \begin{center}
  \begin{tabular}{lccc c ccc}
\hline
\multicolumn{4}{c}{$d_0=10, n_{\mr{train}}=20$}{$d_0=20, n_{\mr{train}}=20$} \hspace{-33ex} \\
\cline{2-4} \cline{6-8} 
Loss    &\hspace{-3ex}$E_{\mr{quad}}$ &\hspace{-3ex} $E_{\mr{QRE}}$ &\hspace{-3ex} $E_{\mr{Stein}}$ & \hspace{-3ex}~ &\hspace{-3ex} $E_{\mr{quad}}$ &\hspace{-3ex} $E_{\mr{QRE}}$ &\hspace{-3ex} $E_{\mr{Stein}}$\\
\hline
$\ell_{\mr{QRE}}$  &\hspace{-3ex} $\vc{9.7\!\times\! 10^{-5}}$  &\hspace{-3ex} $\vc{7\!\times\!10^{-4}}$ &\hspace{-3ex}  $\vc{5.64}$ &\hspace{-3ex}  &\hspace{-3ex}$\vc{1.1\!\times\! 10^{-4}}$ &\hspace{-3ex}$\vc{1.3\!\times \! 10^{-3}}$ &\hspace{-3ex} $\vc{28.86}$\\
$\ell_{\mr{Stein}}$          &\hspace{-3ex} $0.033$    &\hspace{-3ex} $0.28$   &\hspace{-3ex}  $17.75$&\hspace{-3ex}  &$1.07$ &\hspace{-3ex} $8.18$ &\hspace{-3ex} $96.34$\\
$\ell_{\mr{quad}}$     &\hspace{-3ex} $0.043$    &\hspace{-3ex} $0.72$ & \hspace{-3ex} $33.43$ &\hspace{-3ex}  &\hspace{-3ex}$0.061$&\hspace{-3ex}$1.14$ &\hspace{-3ex}$83.66$\\
\hline
\end{tabular}
\vspace{-0ex}
\\
  \begin{tabular}{lccc c ccc}
\hline
\multicolumn{4}{c}{$d_0=10, n_{\mr{train}}=100$}{$d_0=20, n_{\mr{train}}=100$} \hspace{-33ex} \\
\cline{2-4} \cline{6-8} 
Loss    &\hspace{-3.5ex}$E_{\mr{quad}}$ &\hspace{-3.5ex} $E_{\mr{QRE}}$ &\hspace{-3.5ex} $E_{\mr{Stein}}$ & \hspace{-3.5ex}~ &\hspace{-3.5ex} $E_{\mr{quad}}$ &\hspace{-3.5ex} $E_{\mr{QRE}}$ &\hspace{-3ex} $E_{\mr{Stein}}$\\
\hline
$\ell_{\mr{QRE}}$  &\hspace{-3.5ex} $\vc{3.3\!\times\! 10^{-8}}$  &\hspace{-3.5ex} $\vc{5.8\!\times\!10^{-6}}$ &\hspace{-3.5ex}  $\vc{0.31}$ &\hspace{-3.5ex}  &\hspace{-3.5ex}$\vc{6.3\!\times\! 10^{-6}}$ &\hspace{-3.5ex}$\vc{1.1\!\times \! 10^{-4}}$ &\hspace{-3.5ex} $\vc{19.24}$\\
$\ell_{\mr{Stein}}$          &\hspace{-3.5ex} $8.9\!\times\!10^{-4}$    &\hspace{-3.5ex} $0.016$   &\hspace{-3.5ex}  $11.18$&\hspace{-3.5ex}  &$1.2$ &\hspace{-3.5ex} $8.12$ &\hspace{-3.5ex} $85.73$\\
$\ell_{\mr{quad}}$     &\hspace{-3.5ex} $0.037$    &\hspace{-3.5ex} $0.669$ & \hspace{-3.5ex} $38.7$ &\hspace{-3.5ex}  &\hspace{-3.5ex}$0.060$&\hspace{-3.5ex}$1.15$ &\hspace{-3.5ex}$87.62$\\
\hline
\end{tabular}
\vspace{-4.5ex}
\end{center}
    \label{tb:ex1}
\end{table}
\vspace*{-1ex}
\subsubsection{Shallow vs Deep SPD Matrix Learning}
\label{sec:shvsde}
The design of the mMLP model in \eqref{eq:mlpa} enables a mechanism for deep SPD matrix learning by satisfying the SPD constraint across all input, hidden and output layers. The simpler approach would be to consider the standard MLP architecture across input and hidden layers but make use of the activation matrix functions only at the output layer to meet the SPD requirement:
\begin{align}
\label{eq:mlps}
\begin{split}
&
\widehat{\vc Y} = \mc H(\vc Z_0), \quad
\vc Z_0 = \vc W_0 \vc h_1 (\vc W_0 \bc 1)^\top + \vc B_0, \\
&
\vc h_l = \mathfrak h(\vc z_l), \quad
\vc z_l = \vc W_l \vc h_{l+1} + \vc b_l,
\\
&
\vc h_{j+1} = \mathfrak h(\vc z_{j+1}), \quad
\vc z_{j+1} = \vc w_{j+1} \mr{vec}\vc X  + \vc b_{j+1}.
\end{split}
\end{align}
This amounts to a shallow design in the sense that it does not enable a mechanism for preserving the SPD constraint across all layers during the learning.

The design in \eqref{eq:mlpa} allows nonlinearities to pass through layers via activation function matrices which impose the SPD constraint, whereas in the shallow design, nonlinearities are propagated across layers via activation functions without imposing any constraints. 
Our hypothesis is that the former has advantage over the latter in that it captures complex dependencies which are important for the SPD matrix learning. Below we present a numerical example which highlights the importance of preserving the SPD constraint across all layers when learning the SPD matrix.
 
\vspace{-2ex}
\paragraph{Example~2.}
Consider a similar experiment as in Example~1 for the case of ${n_{\mr{train}}=20}$ and output dimensions ${d_0=\{10, 20\}}$ (using a different random seed from Example~1). We directly compare the performance of \eqref{eq:mlpa} against \eqref{eq:mlps} under different number of hidden layers ${j=\{2, 4, 6\}}$ (20 units per layer). The shallow design \eqref{eq:mlps} uses the hyperbolic tangent as the activation function $\mathfrak h(\cdot)$. The same choice of the activation matrix function $\mc H(\cdot)$, given by \eqref{eq:msk}, is used for both models. We use $\ell_{\mr{QRE}}$ as the choice of the loss function (refer to Appendix~\ref{app:example2} for further details). 
The performance is evaluated in terms of $E_{\mr{QRE}}$. 

Table~\ref{tb:ex2} summarizes the results of the evaluation. Although the shallow design \eqref{eq:mlps} performs relatively well, it underperforms in comparison to \eqref{eq:mlpa}. Given the limited number of training samples, arbitrary increasing the number of layers may not be necessarily advantageous, which is the case for both models. However, in this regard, the design in \eqref{eq:mlps} is fairly more sensitive. 
\vspace*{-1ex}
\subsection{Experiments using VAE Models}
 \label{sec:exp_vae}
 \begin{table}[t]
 \vspace{-3ex}
  \tiny
  \caption{SPD matrix learning using the  mMLP (Example~2).}
  \vspace{-2ex}
  \begin{center}
  \begin{tabular}{lccc c ccc}
\hline
\multicolumn{4}{c}{$d_0=10$}{$d_0=20$} \hspace{-24ex} \\
\cline{2-4} \cline{6-8} 
Model    &\hspace{-3.5ex}${j=2}$ &\hspace{-3.5ex} ${j=4}$ &\hspace{-3.5ex} ${j=6}$ & \hspace{-3.5ex}~ &\hspace{-3.5ex} ${j=2}$ &\hspace{-3.5ex} ${j=4}$ &\hspace{-3ex} ${j=6}$\\
\hline
\eqref{eq:mlpa} &\hspace{-3.5ex} $\vc {2\!\times\! 10^{-4}}$  &\hspace{-3.5ex} $\vc{1\!\times\!10^{-5}}$ &\hspace{-3.5ex}  $\vc{4\!\times\! 10^{-5}}$ &\hspace{-3.5ex}  &\hspace{-3.5ex}$\vc{5\!\times\! 10^{-3}}$ &\hspace{-3.5ex}$\vc{3\!\times \! 10^{-4}}$ &\hspace{-3.5ex} $\vc{6\times 10^{-4}}$\\
\eqref{eq:mlps}         &\hspace{-3.5ex} $4\!\times\!10^{-3}$    &\hspace{-3.5ex} $5\!\times\!10^{-3}$   &\hspace{-3.5ex}  $4\!\times\!10^{-2}$&\hspace{-3.5ex}  &\hspace{-3.5ex}$8\!\times\!10^{-2}$ &\hspace{-3.5ex} $6\!\times\!10^{-2}$ &\hspace{-3.5ex} $7\!\times\!10^{-1}$\\
\hline
\end{tabular}
\vspace{-5ex}
\end{center}
  \label{tb:ex2}
\end{table}
 \begin{table*}[h]
 \vspace*{-1ex}
 \begin{center}
 \vspace{-2ex}
\caption{Output activation matrix functions and activation functions for the models in Table~\ref{tb:abb}}
  \tiny
\begin{tabular}{lcccccc}
\hline
\multicolumn{5}{c}{$\qquad \qquad $ Generative Network} \\
\cline{2-6}
Model    & $\bc \mu_p$   &  $\log \eta_p$ & $\alpha_p$ & $\beta_p$ & $\bc \Omega_p$ ~      \\
\hline
${\mc N_\mr{d}\mc N_\mr{d}}$     & $\mr{linear}()$   &  $\mr{linear}()$ & -- & -- & * ~      \\
${\mc N_\mr{d}\mc N_\mr{f}}$     & $\mr{linear}()$   &  $\mr{linear}()$ & -- & -- & * ~      \\
${\mc N_\mr{f}\mc N_\mr{d}}$     & $\mr{linear}()$   &  $\mr{linear}()$ & -- & -- & ** ~      \\
${\mc N_\mr{f}\mc N_\mr{f}}$     & $\mr{linear}()$   &  $\mr{linear}()$ & -- & -- & ** ~      \\
${\mc E_\mr{f}\mc N_\mr{f}}$     & $\mr{linear}()$   &  $\mr{linear}()$ & 0.5+$\mr{sigmoid}()$ & 0.5+$\mr{sigmoid}()$ & ** ~      \\
${\mc E_\mr{f}\mc E_\mr{f}}$     & $\mr{linear}()$   &  $\mr{linear}()$ & 0.5+$\mr{sigmoid}()$ & 0.5+$\mr{sigmoid}()$ & ** ~      \\
\hline
\end{tabular}
\hspace*{-1ex}
\begin{tabular}{lcccccc}
\hline
\multicolumn{5}{c}{$\qquad \qquad $ Recognition Network} \\
\cline{2-6}
~    & $\bc \mu_q$   &  $\log \eta_q$ & $\alpha_q$ & $\beta_q$ & $\bc \Omega_q$ ~      \\
\hline
~     & $\mr{linear}()$   &  $\mr{linear}()$ & -- & -- & * ~      \\
& $\mr{linear}()$   &  $\mr{linear}()$ & -- & -- & ** ~      \\
& $\mr{linear}()$   &  $\mr{linear}()$ & -- & -- & * ~      \\
& $\mr{linear}()$   &  $\mr{linear}()$ & -- & -- & ** ~      \\
& $\mr{linear}()$   &  $\mr{linear}()$ & -- & -- & ** ~      \\
& $\mr{linear}()$   &  $\mr{linear}()$ & 0.5+$\mr{sigmoid}()$ & 0.5+$\mr{sigmoid}()$ & ** ~      \\
\hline
\end{tabular}\\
\vspace*{-3ex}
\begin{itemize}
\item[*] $\mc H(\vc Z) = \frac{\mc K(\vc Z)}{\mr{tr}(\mc K(\vc Z))}$, where $[\mc K(\vc Z)]_{i,i}=\kappa(\vc z_i, \vc z_i)$ and $[\mc K(\vc Z)]_{i,j}=0, \ \forall i\neq j$. The kernel function $\kappa(\cdot, \cdot)$ is given by \eqref{eq:msk}.\\ \vspace*{-2ex}
\item[**] $\mc H(\vc Z) = \frac{\mc K(\vc Z)}{\mr{tr}(\mc K(\vc Z))}$, where $[\mc K(\vc Z)]_{i,j}=\kappa(\vc z_i, \vc z_j), \ \forall i, j$. The kernel function $\kappa(\cdot, \cdot)$ is given by \eqref{eq:msk}.
\end{itemize}
\vspace*{-5ex}
\label{tb:model_details}
\end{center}
\end{table*}
\begin{table*}[t!]
\vspace{-3ex}
\begin{center}
\caption{Performance evaluation on the Frey Face dataset.}
\setlength{\tabcolsep}{6pt}
\renewcommand{\arraystretch}{1.2}
\tiny
\begin{tabular}{l ccc}
\hline
\multicolumn{2}{c}{${\mc N_\mr{d}\mc N_\mr{d}}$} \hspace{-22ex} \\
\cline{2-3}
$k\quad$    & LL & KLD \\
\hline
$5$ &   $-66.3$  &  $9\times 10^{-6}$  \\
$8$ &  $-64.5$  &  $2\times 10^{-5}$  \\
\end{tabular}
\hspace{-1.5ex}
\vline
\hspace{-1.5ex}
\begin{tabular}{ccc}
\hline
\multicolumn{2}{c}{${\mc N_\mr{d}\mc N_\mr{f}}$} \hspace{0ex} \\
\cline{1-2}
 LL & KLD \\
\hline
  $-66.4$ & $1\times 10^{-6}$    \\
  $-64.2$ & $1\times 10^{-4}$      \\ 
\end{tabular}
\hspace{-1.5ex}
\vline
\hspace{-1.5ex}
\begin{tabular}{ccc}
\hline
\multicolumn{2}{c}{${\mc N_\mr{f}\mc N_\mr{d}}$} \hspace{0ex} \\
\cline{1-2}
 LL & KLD \\
\hline
  $-65.3$ & $3\times 10^{-3}$      \\
  $-63.9$ & $5\times 10^{-3}$      \\
\end{tabular}
\hspace{-1.5ex}
\vline
\begin{tabular}{ccc}
\hline
\multicolumn{2}{c}{${\mc N_\mr{f}\mc N_\mr{f}}$} \hspace{0ex} \\
\cline{1-2}
 LL & KLD \\
\hline
  $-64.4$ & $0.65$      \\
  $-63.1$ & $0.46$      \\
\end{tabular}
\hspace{-1.5ex}
\vline
\hspace{-1.5ex}
\begin{tabular}{ccc}
\hline
\multicolumn{2}{c}{${\mc E_\mr{f}\mc N_\mr{f}}$} \hspace{0ex} \\
\cline{1-2}
 LL & KLD \\
\hline
 $-63.5$ & $1.04$     \\
  $-62.3$ & $0.71$      \\
\end{tabular}
\hspace{-1.5ex}
\vline
\hspace{-1.5ex}
\begin{tabular}{ccc}
\hline
\multicolumn{2}{c}{${\mc E_\mr{f}\mc E_\mr{f}}$} \hspace{0ex} \\
\cline{1-2}
 LL & KLD \\
\hline
  $-64.0$ & $56.2$      \\
  $-62.8$ & $40.6$      \\
\end{tabular}
\vspace*{-5ex}
\label{tb:frey_vae}
\end{center}
\end{table*}
 We train generative models of images from the Frey Face dataset\footnote{Available at: https://cs.nyu.edu/~roweis/data.html}. The dataset consists of images of size ${20\times 28}$ taken from sequential frames of a video which can be treated as continuous real space. The size of images poses a computational challenge for those model variants with the dense dispersion matrices in their generative networks since we would need to learn SPD matrices of size ${560\times 560}$. As the primary goal of this experiment is to provide insights into the questions raised in Section~\ref{sec:back_vae}, for the computational reasons (see Section~\ref{sec:diss}), we extract the first $10$ principal components of the input images and carry out the analysis on the resulting principal components. This would allow us to evaluate all model variants, listed in Table~\ref{tb:abb}, within the same pipeline. The only difference is the parametric choice for the generative and the recognition networks. In terms of the degree of flexibility, the following is true:
 \begin{align}
 \label{eq:model_flex}
{\mc N_\mr{d}\mc N_\mr{d}} < {\mc N_\mr{d}\mc N_\mr{f}} \approx {\mc N_\mr{f}\mc N_\mr{d}} < {\mc N_\mr{f}\mc N_\mr{f}} < {\mc E_\mr{f}\mc N_\mr{f}} < {\mc E_\mr{f}\mc E_\mr{f}},
 \end{align}
where ${{\mc N_\mr{d}\mc N_\mr{f}} \approx {\mc N_\mr{f}\mc N_\mr{d}} }$ since, at this point, it is not obvious which model is the more flexible one.

All models use mMLPs with 3 layers and $30$ units per layer. The same choices of activation matrix functions, via the Mercer sigmoid kernel function \eqref{eq:msk}, and activation functions, via the hyperbolic tangent function, are used across all layers for all models. However, the output activation matrix functions and the output activation functions are model dependent. These are summarized in Table~\ref{tb:model_details}.

Data are divided into a training set (1000 samples) and a test set (965 samples). The VAE models are trained on the training set and evaluated on the test set in terms of the log-likelihood (LL) and the KLD between the posterior and the prior. 
The experiment is repeated for different latent variable dimensions, ${k=\{5, 8\}}$. 
Table~\ref{tb:frey_vae} summarizes the results of the evaluation.

 The first observation is that the model variants with full covariance matrices at both their generative and recognition networks (${\mc N_\mr f \mc N_{\mr f}}, {\mc E_{\mr f} \mc N_{\mr f}}, {\mc E_{\mr f} \mc E_{\mr f}}$) outperform the ones that impose the diagonality constraint on either the generative network or the recognition network, in terms of the log-likelihood scores. Both ${\mc E_{\mr f} \mc N_{\mr f}}$ and ${\mc E_{\mr f} \mc E_{\mr f}}$ consistently perform better than ${\mc N_\mr f \mc N_{\mr f}}$ which may further suggest that increasing the flexibility of the generative network, in this case by relaxing the Gaussian assumption, might be advantageous. Between the two model variants that use the mPE distribution in their generative networks, namely ${\mc E_{\mr f} \mc N_{\mr f}}$ and ${\mc E_{\mr f} \mc E_{\mr f}}$, the model variant ${\mc E_{\mr f} \mc N_{\mr f}}$ achieved the highest log-likelihood score. This might seem counterintuitive since, as shown in \eqref{eq:model_flex}, ${\mc E_{\mr f} \mc E_{\mr f}}$ allows higher flexibility on the recognition model in comparison to ${\mc E_{\mr f} \mc N_{\mr f}}$---in other words, one might expect this additional flexibility to be translated directly into higher log-likelihood scores. However, this could be explained by the fact that ${\mc E_{\mr f} \mc E_{\mr f}}$ uses the estimator \eqref{eq:est_mpe} which has potentially higher variance than the estimator \eqref{eq:est_n} used by ${\mc E_{\mr f} \mc N_{\mr f}}$ (recall the discussion in Section~\ref{sec:vae_model_const}).

 The next observation is that, in terms of the KLD between the posterior and the prior, the more flexible model variants score higher. If the KLD is close to zero, one might argue that the VAE model has partly failed to code any information into the latent variables. Thus, KLD scores greater than zero might be indeed desirable. The model variant ${\mc E_{\mr f} \mc E_{\mr f}}$ that allows the highest degree of flexibility on the recognition network has the highest KLD. 
Finally, we generated random samples from the generative network of each model which are shown in Figures~\ref{fig:5} and \ref{fig:8}, available in the supplemental material.
\vspace*{-1ex}
\section{Limitations and Future Work}
\label{sec:diss}
\vspace*{-1ex}
The main limitation of the mMLP has to do with scalability to higher dimensions. The complexity  associated with computing the $\alpha$-derivative of the von Neumann loss function~\eqref{eq:loss_qre} at the output layer is $\mc O(d_0^3)$. Taking the symmetric nature of the SPD matrices into account, the computational complexity at the hidden layer $l$ reduces to $\mc O(d_l^2)$. 

The current implementation of the matrix backpropagation involves multiple applications of the Kronecker product. Although it facilitates the implementation, we would need access to the full Jacobian matrices (${d_l^2\times d_l^2}$). However, these matrices are in fact available in the form of sparse block matrices, which means that it is possible to implement a  memory-efficient computation of the tensor products without the need to actually have access to the full matrices. Future work is needed in this direction.

Within the mMLP, the choice of loss function can be consequential (recall  Example~1). In this regard, we showed the effectiveness of the von Neumann loss function \eqref{eq:loss_qre}. However, in connection to the VAE, we used the ELBO as the objective function which includes a KLD (relative entropy) term. It would be interesting to alter the VAE objective function such that it takes advantage of the von Neumann entropy (the quantum relative entropy) in its formulation.

Another interesting direction for future work is to investigate other applications of the proposed mMLP model. One possibility is in the context of the heteroscedastic multivariate regression, and we believe that there are many other cases in which the mMLP can prove to be useful.
\vspace{-1ex}
\section{Discussion}
\vspace{-1ex}

We introduced a tool to learn SPD matrices, referred to as the matrix multilayer perceptron (mMLP). The mMLP takes the non-Euclidean geometry of the underlying SPD manifolds into account by making use of the von Neumann divergence as the choice of the SPD manifold metric. One key aspect of the mMLP is that it preserves the SPD constraint across all layers by exploiting PD kernel functions and a backpropagation algorithm that respects the inherent SPD nature of the matrices.

We presented an application of the mMLP in connection to the VAE. 
Integrating the mMLP in the VAE allowed us to consider parametric families of distributions with dense covariance matrices. Two candidates were discussed: the Gaussian distribution with a dense covariance matrix and its generalization to the mPE distribution.
Based on these choices, we constructed six model alternatives with various degrees of flexibility. 
Our results support the importance of increasing the flexibility of the VAE's recognition network, which is in line with the current  understanding in the VAE. However, we also found that it is \textit{just as important} to increase the flexibility of the generative network. Importantly, we found no signs of overfitting in doing so: The two model variants that achieved the highest likelihood and the largest KLD scores were indeed among the first two most flexible ones. 




\section*{Acknowledgements}
This research is financially supported by The Knut and Alice Wallenberg Foundation (J. Taghia, contract number: KAW2014.0392), by the project \emph{Learning flexible models for nonlinear dynamics} (T.~B.~Sch{\"o}n, contract number: 2017-03807), funded by the Swedish Research Council and by the Swedish Foundation for Strategic Research (SSF) via the project \emph{ASSEMBLE} (T.~B.~Sch{\"o}n, contract number: RIT15-0012), by the project 
\emph{Learning of Large-Scale Probabilistic Dynamical Models} (F. Lindsten, contract 
number: 2016-04278) funded by the Swedish Research Council 
and by the Swedish Foundation for Strategic Research via the project
\emph{Probabilistic Modeling and Inference for Machine Learning} 
(F. Lindsten, contract number: ICA16-0015). We thank Carl Andersson for useful discussions.
\bibliography{ref}
\bibliographystyle{icml2019}

\clearpage
\onecolumn
\appendix
\renewcommand\thefigure{\thesection.\arabic{figure}}  
\numberwithin{figure}{section}
\numberwithin{equation}{section}
\renewcommand\thetable{\thesection.\arabic{table}}  
\numberwithin{table}{section}

\section{Matrix notations}
\label{app:notation} 
We use $\top$ for the transpose operator, $\msf{tr}(\cdot)$ for the trace operator, and $\msf{det}(\cdot)$ for the matrix determinant.
 The symmetric part of a square matrix $\vc B$ is denoted by ${\sym (\vc B)={(\vc B+\vc B^\top)}/{2}}$. 
 The Kronecker product is denoted by $\otimes$, the Hadamard product by $\circ$, and the dot product by $\odot$.
We use the vec-operator for \emph{column-by-column} stacking of a matrix $\vc A$, shown as ${\msf{vec}\vc A\equiv \msf{vec}(\vc A)}$. Let $\vc A$ be an ${m\times n}$ matrix, the operator ${\msf P_{(m, n)}}$ will then rearrange $\msf{vec}\vc A$ to its matrix form: ${\vc A= \msf P_{(m, n)}(\msf{vec}\vc A)}$. For the ${m\times n}$ dimensional matrix $\vc A$, the commutation matrix is shown as $\bc K_{(m,n)}$ which is the ${mn\times mn}$ matrix that transforms ${\msf {vec}\vc A}$ into ${\msf {vec}\vc A^\top}$ as: ${\bc K_{(m,n)}\msf{vec}\vc A= \msf{vec}\vc A^\top}$. An ${m\times m}$ identity matrix is shown as $\bc I_{m}$.
If ${\mc T(\vc X):\mbb R\p{d\times d}\rightarrow \mbb R}$ is a real-valued function on matrices, then ${\vc \nabla_{\vc X} \mc T(\vc X)}$ denotes the gradient with respect to the matrix $\vc X$, ${\vc \nabla_{\vc X} \mc T(\vc X) = \big[ \frac{\partial \mc T}{\partial \mr x_{ij}}  \big]_{i,j=1:d}}$.
The \emph{matrix logarithm} and the \emph{matrix exponential} are written as $\blog \vc A$ and $\bexp\vc A$, respectively. The matrix exponential in the case of symmetric matrices can be expressed using the eigenvalue decomposition as ${\bexp \vc A=\vc V(\bexp\vc \Lambda) \vc V^\top}$, where $\vc V$ is an orthonormal matrix of eigenvectors and $\vc \Lambda$ is a diagonal matrix with the eigenvalues on the diagonal. The matrix logarithm is the inverse of the matrix exponential if it exists. If $\vc A$ is symmetric and strictly positive definite (PD) it is computed using ${\blog\vc A= \vc V (\log \vc\Lambda)\vc V^\top}$, where ${(\log \vc \Lambda)_{i,i}} = \log \Lambda_{i,i}$.

\section{The $\alpha$-derivative: Definition and properties}
\label{app:alpha}
\paragraph*{Definition.}
Let $\vc F$ be an $m\times n$ matrix function of an $n\times q$ matrix of variables $\vc X$. The $\alpha$-derivative of $\vc F(\vc X)$ is defined as \citep[Definition~2]{Magnus2010}
\begin{align}
 {\msf D_{\vc X}\vc F :=\frac{\partial\ \msf{vec}\vc F(\vc X)}{\partial \ (\msf{vec}\vc X)^\top}},
\end{align}
where $\msf D_{\vc X}\vc F$ is an $mp \times nq$ matrix which contains all the partial derivatives such that each row contains the partial derivatives of one function with respect to all variables, and each column contains the partial derivatives of all functions with respect to one variable.
\paragraph*{Product rule.}
Let $\vc F$ ${(m\times p)}$ and $\vc G $ ${(p\times r)}$ be functions of $\vc X$ ${(n\times q)}$. Then the product rule for the $\alpha$-derivative is given by \citep{Magnus2010}
\begin{align}
\label{eq:product_alpha}
\msf D_{\vc X} (\vc F\vc G) = (\vc G^\top \otimes \bc I_m) \msf D_{\vc X} \vc F+ (\bc I_r \otimes \vc F)\msf D_{\vc X} \vc G.
\end{align}
\paragraph*{Chain rule.} Let $\vc F$ (${m\times p}$) be differentiable at $\vc X$ (${n\times q}$), and $\vc G$ (${l\times r}$) be differentiable at ${\vc Y=\vc F(\vc X)}$, then the composite function ${\vc H(\vc X) = \vc G(\vc F(\vc X))}$ is differentiable at $\vc X$, and
\begin{align}
\label{eq:chain_alpha}
\msf D_{\vc X} \vc H = \msf D_{\vc Y}\vc G \msf D_{\vc X}\vc F,
\end{align}
which expresses the chain rule for the $\alpha$-derivative \citep{Magnus2010}.

\section{The Stein divergence }
\label{app:Stein} 
An important choice of function in \eqref{eq:bergman} is provided by ${\mc {F}(\vc X)=-\log \mr{det}}(\vc X)$, under which the Stein divergence \citep{Stein1956}, commonly known as the LogDet Divergence \citep{Kulis2009}, is obtained as 
\begin{align}
\label{eq:div}
\!\!\!\!\!\Delta_{\mr {Stein}}(\vc X || \widetilde{\vc X})\!= \!\mr{tr}(\widetilde{\vc X}^{-1} (\vc X - \widetilde{\vc X})) \!- \!\log \mr{det}(\vc X\widetilde{\vc X}^{-1}). \hspace{-2ex}
\end{align}
\subsection{The symmetrized Stein divergence}
The symmetrized Stein divergence is defined as
\begin{align}
\label{eq:loss_stein}
\ell_{\mr{Stein}}(\widehat{\vc Y}, \vc Y)  := \Delta_{\mr {Stein}}^{\mr{sym}} (\yh , \vc Y)
=  \frac{1}{2}(\Delta_{\mr {Stein}} (\vc Y|| \bar{\vc Y}) + \Delta_{\mr {Stein}} (\yh|| \bar{\vc Y})),
\end{align}
where ${\bar{\vc Y} = {(\vc Y + \yh)}/{2}}$, and $\Delta_{\mr {Stein}}$ is given in \eqref{eq:div}. The symmetrization follows from the Jensen-Bregman divergence \citep[refer to][]{Nielsen2011}. 
\subsection{The $\alpha$-derivative of the symmetrized Stein divergence}
\label{sec:der_loss2}
The $\alpha$-derivative of the symmetrized Stein divergence \eqref{eq:loss_stein} for ${\{\vc Y, \yh\} \in \mbb R^{d_0\times d_0}}$, can be expressed as (keeping terms that only depend on $\yh$)
\begin{multline}
\msf D_{\yh} \ell_{\mr{Stein}}(\yh, {\vc Y}) 
= \msf D_{\yh} \mr{tr}((\vc Y + \yh)^{-1} \vc Y) - \frac{1}{2}\msf D_{\yh}\log| \vc Y (\vc Y + \yh)^{-1} | + \msf D_{\yh} \mr{tr}((\vc Y + \yh)^{-1} \yh) - \frac{1}{2}\msf D_{\yh}\log| \yh (\vc Y + \yh)^{-1} | \\ = \underbrace{\msf D_{\yh} \mr{tr}((\vc Y + \yh)^{-1} (\vc Y+\yh)}_{=0} - \frac{1}{2} (\msf D_{\yh}\log| \vc Y (\vc Y + \yh)^{-1} | +\msf D_{\yh}\log| \yh (\vc Y + \yh)^{-1} | ),
\end{multline}
where the remaining terms are computed via the repeated use of the $\alpha$-derivative's product and chain rules as
\begin{align}
&\msf D_{\yh}\log| \vc Y (\vc Y + \yh)^{-1} | =  (\mr{vec}(\vc Y (\vc Y + \yh)^{-1})^{-\top})^{\top} \msf D_{\yh} \vc Y (\vc Y + \yh)^{-1}, \\
& 
\msf D_{\yh}\log| \yh (\vc Y + \yh)^{-1} | = -(\mr{vec}(\yh (\vc Y + \yh)^{-1})^{-\top})^{\top} \msf D_{\yh} \vc Y (\vc Y + \yh)^{-1},
\end{align}
where 
\begin{align}
\msf D_{\yh} \vc Y (\vc Y + \yh)^{-1} =-(\bc I_{d_0} \otimes \vc Y)((\vc Y + \yh)^{-\top} \otimes (\vc Y + \yh)^{-1}).
\end{align}
\section{The $\alpha$-derivative of the symmetrized von Neumann divergence}
\label{sec:der_loss}
For the loss function defined in \eqref{eq:loss_qre}, using the $\alpha$-derivative's product rule and chain rule, we obtain
\begin{align}
\label{eq:dloss}
\begin{split}
\msf D_{\yh} \ell 
&= \frac{1}{2} \msf D_{\yh}\msf {tr}((\yh-\vc Y)\blog \yh)  - \frac{1}{2} \msf D_{\yh} \msf{tr} (\yh \blog \vc Y) ,
\end{split}
\end{align}
where the above two terms are computed using 
\begin{align}
\label{eq:loss_ja}
&\msf D_{\yh}(\msf {tr}((\yh-\vc Y)\blog \yh)) = \underbrace{(\msf{vec} (\blog \yh)^\top)^\top}_{1\times d_0^2 } +
\underbrace{\begin{pmatrix} 
\msf{vec} (\yh^\top - \vc Y^\top) \odot \msf{vec}( \frac{\partial}{\partial \widehat{\mr Y}_{11} } \blog \yh) \\ \msf{vec} (\yh^\top - \vc Y^\top) \odot \msf{vec}( \frac{\partial}{\partial \widehat{\mr Y}_{21} } \blog \yh)  \\
\vdots \\
\msf{vec} (\yh^\top - \vc Y^\top) \odot \msf{vec}(\frac{\partial}{\partial \widehat{\mr Y}_{d_0d_0} } \blog \yh) 
\end{pmatrix}^\top}_{1\times d_0^2 },
\\ 
&\msf D_{\yh}( \msf{tr} (\yh \blog \vc Y) )  = \underbrace{(\msf{vec}  (\blog \vc Y)^\top)^\top}_{1\times d_0^2}.
\end{align}

Refer to Appendix~\ref{app:notation} for a summary of the notation. 
The remaining part in the computation of \eqref{eq:dloss} is to evaluate ${\frac{\partial}{\partial \widehat{\mr Y}_{ij} }\blog \yh }$ for all ${i,j\in\{1, \ldots,d_0\}}$, which involves taking derivatives through the eigendecomposition. In the following, we take a similar approach as in \citet{Papadopoulo2000} to compute the necessary partial derivatives.

Let $\yh= \vc \Upsilon\mr{diag}(\lambda_1, \ldots, \lambda_{d_0})\vc \Upsilon^\top$ be the notion of our eigendecomposition. We can write 
\begin{align}
\label{eq:jacob1}
\begin{split}
{\frac{\partial}{\partial \widehat{\mr Y}_{ij} }\blog \yh }&= \frac{\partial}{\partial \widehat{\mr Y}_{ij} }
 \vc \Upsilon  \vc \Lambda \vc \Upsilon^\top, \quad \text{where\ \ }\vc \Lambda=\mr{diag}(\log \lambda_1, \ldots, \log \lambda_{d_0}),
\\ 
&=
\frac{\partial \vc \Upsilon }{\partial \widehat{\mr Y}_{ij}} \vc\Lambda\vc \Upsilon^\top +
\vc \Upsilon \frac{\partial \vc \Lambda }{\partial \widehat{\mr Y}_{ij}} \vc \Upsilon^\top + 
\vc \Upsilon \vc\Lambda \frac{\partial \vc \Upsilon^\top }{\partial \widehat{\mr Y}_{ij}}.
\end{split}
\end{align}
By multiplying \eqref{eq:jacob1} from left and right by $\vc \Upsilon^\top$ and $\vc \Upsilon$ respectively, we obtain:
\begin{align}
\label{eq:jacob2}
\begin{split}
\vc \Upsilon^\top \  \frac{\partial}{\partial \widehat{\mr Y}_{ij} }\blog \yh \ \vc \Upsilon &=
\vc \Upsilon^\top \frac{\partial \vc \Upsilon }{\partial \widehat{\mr Y}_{ij}} \vc\Lambda+
 \frac{\partial \vc \Lambda }{\partial \widehat{\mr Y}_{ij}} + 
\vc\Lambda \frac{\partial \vc \Upsilon^\top }{\partial \widehat{\mr Y}_{ij}}\vc \Upsilon
\\
&=
 \vc \Xi_{ij}(\vc \Upsilon) \vc \Lambda +  \frac{\partial \vc \Lambda }{\partial \widehat{\mr Y}_{ij}} - \vc \Lambda {\vc \Xi_{ij}(\vc \Upsilon)},
 \end{split}
\end{align}
where we have defined ${\vc \Xi_{ij}(\vc \Upsilon) = \vc \Upsilon^\top \frac{\partial }{\partial \widehat{\mr Y}_{ij}}\vc \Upsilon }$ and used the fact that ${\vc \Xi_{ij}(\vc \Upsilon)}$ is an antisymmetric matrix, ${\vc \Xi_{ij}(\vc \Upsilon) + \vc \Xi_{ij}^\top (\vc \Upsilon)= \vc 0}$, which in turn follows from the fact that $\vc \Upsilon $ is an orthonormal matrix,
\begin{align}
 {\vc \Upsilon ^\top \vc \Upsilon = \bc I_{d_0} \Rightarrow \frac{\partial \vc \Upsilon^\top }{\partial \widehat{\mr Y}_{ij}} \vc \Upsilon + \vc \Upsilon^\top {\frac{\partial \vc \Upsilon }{\partial \widehat{\mr Y}_{ij}} } = \vc \Xi_{ij}^\top(\vc \Upsilon) + \vc \Xi_{ij}(\vc \Upsilon) = \vc 0}.
 \end{align}
Taking the antisymmetric property of $\vc \Xi_{ij}(\vc \Upsilon)$ into account in \eqref{eq:jacob2}, we obtain
\begin{align}
\label{eq:key1}
&\frac{\partial}{\partial \widehat{\mr Y}_{ij}} \log \lambda_{k} = \Upsilon_{ik} \Upsilon_{jk},
\\ \label{eq:jacob3}
& \vc \Xi_{ij}(\Upsilon_{kl}) = \frac{\Upsilon_{ik} \Upsilon_{jl} + \Upsilon_{il} \Upsilon_{jk}}{2(\log \lambda_l - \log \lambda_k )}, \quad \forall l\neq k.
\end{align}
It is notable that by construction, we do not have repeating eigenvalues, that is $\lambda_k\neq \lambda_l, \ \forall k\neq l$, so there exists a unique solution to \eqref{eq:jacob3}. 
Once $\vc \Xi_{ij}(\vc \Upsilon)$ is computed, it follows that 
\begin{align}
\label{eq:key2}
\frac{\partial \vc \Upsilon }{\partial \widehat{\mr Y}_{ij}} = \vc \Upsilon \vc \Xi_{ij}(\vc \Upsilon), \qquad \frac{\partial \vc \Upsilon^\top }{\partial \widehat{\mr Y}_{ij}} = -\vc \Xi_{ij}(\vc \Upsilon) \vc \Upsilon^\top .
\end{align}

In summary, the necessary partial derivatives for computing \eqref{eq:jacob1} is given by \eqref{eq:key1} and \eqref{eq:key2}. Once \eqref{eq:jacob1} is computed for all $i,j$, we can evaluate \eqref{eq:loss_ja} and ultimately evalaute \eqref{eq:dloss}.

\section{The basic case of the mMLP}
\label{app:basic_mlp}
\subsection{Feedforward step}
\label{par:forward}
At the feedforward computation, we compute and store the latent outputs $\yh$, $\vc H_l$ for all ${l\in\{j+1, \ldots, 1\}}$ using the current setting of the parameters, which are $\vc W_l$, and $\vc B_l$ computed from the learning step, Appendix~\ref{sec:learning}.
\subsection{Backpropagation step}
\label{par:backward}
We first summarize the necessary $\alpha$-derivatives for the backpropagation, and then write down the backpropagation procedure accordingly.
\paragraph{Derivatives required for backpropagation.}
\label{sec:derb1}
The derivative of the activation matrix function depends on the specific choice of kernel function, and in general it is computed readily from the definition of $\alpha$-derivative,
\begin{align}
\msf D_{\vc Z_l}\mc H(\vc Z_l) &:=\frac{\partial\ \msf{vec}\mc H(\vc Z_l)}{\partial \ (\msf{vec}\vc Z_l)^\top}, \qquad l\in\{0, \ldots, j+1\}.
\end{align}
For our specific choice of activation function, the Mercer Sigmoid kernel~\eqref{eq:msk}, it is computed in Appendix~\ref{app:mercer}.

Via repeated use of the product rule of $\alpha$-derivatives \eqref{eq:product_alpha}, we obtain
\begin{align}
&
\msf D_{\vc W_l} \vc Z_l = (\vc W_l\otimes \bc I_{d_l}) (\vc H_{l+1}^\top \otimes \bc I_{d_l}) +  (\bc I_{d_l} \otimes (\vc W_l\vc H_{l+1}))\bc K_{(d_l, d_{l+1})}, \quad l\in\{0, \ldots, j\},
\\
&
\msf D_{\vc W_{j+1}} \vc Z_l = (\vc W_{j+1} \vc 1_{p_1p_2}\otimes \bc I_{d_{j+1}}) ((\mr{vec}\vc X)^\top \otimes \bc I_{d_{j+1}})+(\bc I_{d_{j+1}} \otimes (\vc W_{j+1}\mr{vec}\vc X))(\bc I_{d_{j+1}} \otimes \vc 1_{p_1p_2}^\top)\bc K_{(d_{j+1}, p_1p_2)},
\\
&
\msf D_{\vc H_{l+1}} \vc Z_l = ( \vc W_{l} \otimes \bc I_{d_{l}} )(\bc I_{d_{l+1}}\otimes \vc W_l), \quad l\in\{0, \ldots, j\},
\end{align}
where $\bc K$ is the commutation matrix (refer to Appendix~\ref{app:notation} for a summary of the matrix notation).
\paragraph{Backpropagation.}
In the interest of simple expressions, let ${\vc H_0\equiv \widehat{\vc Y}}$. Backpropagation to the hidden layer ${l}$ is computed recursively using the derivatives computed at the previous layer according to
\begin{align}
&\msf D_{\vc Z_l} \ell = \msf D_{\vc H_l} \ell \msf D_{\vc Z_l} \vc H_l, &\quad \forall {l \in \{0,\ldots, j+1\}},
\\
& \msf D_{\vc W_l} \ell = \msf D_{\vc Z_l} \ell \msf D_{\vc W_l} \vc Z_l , &\quad \forall {l \in \{0,\ldots, j+1\}},
\\
&
\msf D_{\vc H_{l+1}} \ell = \msf D_{\vc Z_l} \ell\msf D_{\vc H_{l+1}} \vc Z_l, &\quad \forall {l \in \{0,\ldots, j\}},\\
&
\msf D_{\vc B_l} \ell =  \msf D_{\vc Z_l} \ell, &\quad \forall {l \in \{0,\ldots, j+1\}}.
\end{align}
\subsection{Learning step}
\label{sec:learning}
Learning involves updating the weights $\vc W_l$ and the biases $\vc B_l$ using derivatives computed during the backpropagation step.
These are updated using derivatives $\msf D_{\vc W_l} \ell$ and $\msf D_{\vc B_l} \ell$ for a given learning rate $\eta$ as
\begin{align}
&\vc W_l \leftarrow \vc W_l- \eta \msf{P}_{(d_l, d_{l+1})}(\msf D_{\vc W_l}\ell),  &\quad \forall l\in\{0,\ldots, j+1\},\\
&
\vc B_l \leftarrow \vc B_l- \eta \msf{P}_{(d_l, d_{l})}(\msf D_{\vc B_l} \ell),  &\quad \forall l\in\{0,\ldots, j+1\},
\end{align}
where $\msf{P}$ is the rearrangement operator introduced in Appendix~\ref{app:notation}.
\section{The general form of the mMLP}
\label{app:gen_mlp}
\subsection{Feedforward step}
The forward path involves computing and storing both $\vc h_l$, $\widehat{\vc y}$ and $\vc H_l$, $\yh$ using the current settings of the parameters, for all ${l\in \{0, \ldots, j+1\}}$.
\subsection{Backpropagation step}
\label{par:backward2}
Most of the necessary derivatives are identical to the ones in Appendix~\ref{sec:derb1}. However, there are some additional derivatives needed which we will discuss in the following. We then write down the backpropagation formula.
\subsubsection{Required derivatives for backpropagation}
The derivative of the activation function depends on the choice of the function, and it is computed using the definition of the $\alpha$-derivative,
\begin{align}
\msf D_{\vc z_l} \mathfrak {h}_l(\vc z_l) = {\frac{\partial\ \mathfrak h(\vc z_l)}{\partial \ (\vc z_l)^\top}}, \quad l\in\{0, \ldots, j+1\}.
\end{align}
The other required derivatives are computed as
\begin{align}
&\msf D_{\vc A_l} \vc z_l = 
{\vc C_l \vc H_l}
,
&\qquad l\in \{0, \ldots, j\},
\\
&\msf D_{\vc A_{j+1}} \vc z_{j+1} = 
{(\vc C_{j+1} \vc H_{j+1})}{(\vc 1_{r_{j+1}}^\top \otimes \bc I_{d_{j+1}})},
\\
&
\msf D_{\vc C_l} \vc z_l = 
{(\vc H_l \vc A_l \vc h_{l+1})^\top \otimes \bc I_{r_l}},
&\qquad l\in \{0, \ldots, j\},
\\
&
\msf D_{\vc C_{j+1}} \vc z_{j+1} = 
{(\vc H_{j+1} \vc A_{j+1} \vc 1_{r_{j+1}})^\top \otimes \bc I_{r_{j+1}}},
\\
&
\msf D_{\vc h_{l+1}} \vc z_{l} = {\vc C_l\vc H_l\vc A_l}, &\qquad l\in \{0, \ldots, j\},
\\
&
\msf D_{\vc H_l}\vc z_l = {((\vc A_l \vc h_{l+1})^\top \otimes \bc I_{r_{l}}}){(\bc I_{d_{l}} \otimes \vc C_{l})}
, &\qquad l\in \{0, \ldots, j\},
\\
&
\msf D_{\vc H_{j+1}}\vc z_{j+1} = {((\vc A_{j+1} \vc 1_{j+1})^\top \otimes \bc I_{r_{j+1}}})
{(\bc I_{d_{j+1}} \otimes \vc C_{j+1})}.
\end{align}
\paragraph*{Backpropagation.}
For simplicity of expressions, let $\vc h_0 \equiv\widehat{\vc y}$ and $\vc H_0 \equiv\widehat{\vc Y}$. The derivatives are recursively computed as
\begin{align}
&
\msf D_{\vc h_0} \ell \equiv \msf D_{\widehat{\vc y}} \ell
\\
&
\msf D_{\vc H_0} \ell \equiv \msf D_{\yh} \ell = \msf D_{\vc z_0} \ell \msf D_{\yh} \vc z_0 + \msf D_{\yh} \ell
&
\\
& \msf D_{\vc z_l} \ell = \msf D_{\vc h_l} \ell \msf D_{\vc z_l} \vc h_l, \qquad &\forall l \in \{0,\ldots, j+1\},
\\
&
\msf D_{\vc h_{l+1}} \ell = \msf D_{\vc z_l} \ell \msf D_{\vc h_{l+1}} \vc z_l, \qquad &\forall l \in \{0,\ldots, j\},
\\
&
\msf D_{\vc Z_l} \ell = \msf D_{\vc H_l} \ell\msf D_{\vc Z_l} \vc H_l, \qquad &\forall l \in \{0,\ldots, j+1\},
\\ 
&
\msf D_{\vc H_{l+1}} \ell = \msf D_{\vc z_{l+1}} \ell \msf D_{\vc H_{l+1}} \vc z_{l+1} + \msf D_{\vc Z_l} \ell \msf D_{\vc H_{l+1}} \vc Z_l, \quad &\forall l \in \{0,\ldots, j\},
\\
&
\msf D_{\vc A_l} \ell = \msf D_{\vc z_l} \ell \msf D_{\vc A_l} \vc z_l, \qquad &\forall l \in \{0,\ldots, j+1\},
\\
&
\msf D_{\vc C_l} \ell = \msf D_{\vc z_l} \ell \msf D_{\vc C_l} \vc z_l, \qquad &\forall l \in \{0,\ldots, j+1\},
\\
&
\msf D_{\vc W_{l}} \ell = \msf D_{\vc Z_l} \ell \msf D_{\vc W_l} \vc Z_l, \quad &\forall l \in \{0,\ldots, j+1\},
\\
&
\msf D_{\vc b_l} \ell = \msf D_{\vc z_{l}} \ell, \quad &\forall l \in \{0,\ldots, j+1\},
\\
&
\msf D_{\vc B_l} \ell = \msf D_{\vc Z_{l}} \ell, \quad &\forall l \in \{0,\ldots, j+1\}.
\end{align}
\subsection{Learning step}
\label{sec:learning_2}
The learning step involves updating the weights and the biases which are computed using derivatives computed from the backpropagation step. Update rules for $\vc W_l$, and $\vc B_l$ are the same as the ones given in Appendix~\ref{sec:learning}. The remaining parameters are learned in a similar fashion,
\begin{align}
&\vc A_l \leftarrow \vc A_l- \eta \msf{P}_{(d_l, r_{l+1})}(\msf D_{\vc A_l}\ell),  &\quad \forall l\in\{0,\ldots, j+1\},
\\
&\vc C_l \leftarrow \vc C_l- \eta \msf{P}_{(r_l, d_l)}(\msf D_{\vc C_l}\ell),  &\quad \forall l\in\{0,\ldots, j+1\},
\\&
\vc b_l \leftarrow \vc b_l- \eta \msf{P}_{(r_{l}, 1)}(\msf D_{\vc b_l}\ell),  &\quad \forall l\in\{0,\ldots, j+1\}.
\end{align}
\section{The $\alpha$-derivative of the Mercer sigmoid kernel}
\label{app:mercer}
The $\alpha$-derivative of the Mercer sigmoid kernel can be computed as
\begin{align}
\msf D_{\vc Z} \mc H =
\begin{pmatrix} 
\cdots & \cdots & \cdots 
\\
\cdots  & \underbrace{\frac{\partial}{\partial \vc z_i} \frac{\kappa_{mn}}{\msf{tr}\mc K }}_{1 \times d_l} & \cdots 
\\
\cdots & \cdots & \cdots 
\end{pmatrix}, \qquad \forall\ i, m, n\in\{1,\ldots, d_l\},
\end{align}
where $\vc z_i$ indicates the $i^{\text{th}}$ column of $\vc Z$,  ${\msf{tr}\mc K\equiv\msf{tr}\mc K (\vc Z)}$, ${\kappa_{mn}\equiv\kappa(\vc z_m, \vc z_n)}$ as defined in \eqref{eq:msk}, and
\begin{align}
\frac{\partial}{\partial \vc z_i} \frac{\kappa_{mn}}{\msf{tr}\mc K } = \left(
\frac{\alpha (\vc 1^\top - \mathfrak f(\vc z_i)\circ \mathfrak f(\vc z_i))}{(\msf {tr} \mc K)^2}\right)\circ\left( \msf{tr} \mc K ~ \mathfrak{f}\left(\frac{\partial}{\partial \vc z_i} (\vc z_m \circ \vc z_n)\right) - 2 \kappa_{mn} \mathfrak f(\vc z_i)\right),
\end{align}
where ${\mathfrak f(\vc z_i) := \mr{tanh}(\alpha \vc z_i - \beta)}$.
% ...
\section{Trace-one multivariate Gaussian distribution}
\subsection{Probability density function}
\label{App:t1gauss}
 For a $d$-dimensional random variable ${\bc \vartheta\in \mbb R^{d}}$, we define the trace-one Gaussian distribution according to
\begin{align}
\!\!\!\!\!\mc N_{\mr{tr1}}\!(\bc \vartheta; \bc \mu, \!\bc \Omega, \!\eta) \!=\! \frac{1}{\mr {det}(2\pi \eta \bc \Omega)^{\frac{1}{2}}} \mr{e}^{-\frac{1}{2} (\bc \vartheta - \bc \mu)^\top \!(\eta \bc \Omega)^{-1}\!(\bc \vartheta - \bc \mu)}\!, \hspace*{-1.4ex}
\end{align}
where ${\bc \mu\in \mbb R^{d}}$ is the mean, ${\eta\in \mbb R^+}$ is the scale parameter, and $\bc \Omega\in\mbb{R}^{d\times d}$ is the trace-one covariance matrix, ${\mr {tr}(\bc \Omega)=1}$. The density admits a known stochastic representation in the form of 
\begin{align}
\label{eq:rep}
\! \! \bc \vartheta \overset{d}= \mc T_{\mc N_{\mr{tr1}}}(\bc \epsilon, \bc\vartheta; \phi)= \bc \mu + \bc \Phi \bc \epsilon, \ \   \bc\epsilon\sim N(\vc 0, \bc I), \quad \eta\bc \Omega = \bc \Phi \bc \Phi^\top,
\end{align}
where $\overset{d}=$ denotes equality in distribution and $\phi$ includes the distribution parameters, i.e., ${\phi=\{\bc \mu, \bc \Omega, \eta\}}$.
\subsection{The $\alpha$-derivatives}
\subsubsection{The $\alpha$-derivatives of the logpdf}
\label{app:der_mg}
The $\alpha$-derivatives of the trace-one Gaussian distribution's log-pdf with respect to its parameters are summarized as
\begin{align}
&\msf D_{\bc \Omega} \log \mc N_{\mr{tr1}}(\bc \vartheta; \bc\mu, \bc \Omega, \eta) = -\frac{1}{2} (\mr{vec} (\bc \Omega^{-1}))^\top - \frac{1}{2} (\msf D_{\bc \Omega} t)^\top,
\\&
\msf D_{\bc \Omega} t = - \mr{vec}( (\eta \bc \Omega)^{-1} (\bc \vartheta - \bc \mu) (\bc \vartheta - \bc \mu)^\top \bc \Omega^{-1}),
\end{align}
\begin{align}
\msf D_{\bc \mu}\log \mc N_{\mr{tr1}}(\bc \vartheta; \bc\mu, \bc \Omega, \eta) = (\bc \vartheta - \bc \mu)(\eta \bc \Omega)^{-1},
\end{align}
\begin{align}
\msf D_{\log \eta}\log \mc N_{\mr{tr1}}(\bc \vartheta; \bc\mu, \bc \Omega, \eta) = -\frac{d}{2} + \frac{1}{2} (\bc \vartheta - \bc \mu)^\top (\eta\bc \Omega)^{-1} (\bc \vartheta - \bc \mu).
\end{align}
\subsubsection{The $\alpha$-derivatives of the stochastic representation}
\label{app:der_mg_sr}
For the stochastic representation of the trace-one Gaussian distribution given in \eqref{eq:rep}, we have
\begin{align}
&\msf D_{\bc \Omega} \bc \vartheta = \bc I_d,
\\
& 
\msf D_{\bc \Omega} \bc \vartheta = \eta (\bc \epsilon^\top \otimes \bc I_d) (\bc \Phi^{-1} \otimes \bc I_d),
\\
&
\msf D_{\log \eta} = \eta (\bc \epsilon^\top \otimes \bc I_d) \mr{vec}(\bc \Omega \Phi^{-\top}).
\end{align}

\section{Trace-one multivariate power exponential (mPE) distribution}
\subsection{Probability density function}
The functional form of the probability density function of the mPE distribution is discussed in Section~\ref{sec:ParFamDist}.
\subsection{Moments}
\label{app:mpe_mom}
Let ${\bc\vartheta\in\mbb R^{d}}$ and ${\bc\vartheta\sim {\mc {E}_{\mr{tr1}}}(\bc\mu, \vc\Omega, \eta, \alpha, \beta)}$. The mPE's mean vector and covariance matrix are computed from:
\begin{align}
\label{eq:meanandcov}
\mathbb{E}[\bc \vartheta] = \bc \mu, \qquad
\mbb{V}[\bc\vartheta] = \alpha\eta \nu(\beta)  \vc \Omega, \qquad  \nu(\beta)\!:=\frac{2^{1/ \beta} \Gamma( \frac{ d + 2 }{2\beta})}{d \Gamma(\frac{d}{2\beta} )},
\end{align}
where ${\Gamma(\cdot)}$ denotes the gamma function.
\subsection{The $\alpha$-derivatives}
\subsubsection{The $\alpha$-derivatives of the log-pdf}
\label{app:der_mpe}
It is straightforward to take derivatives of the mPE's log-pdf using the favorable generalization properties of the $\alpha$-derivative's chain and product rules. These are summarized as:
\begin{align}
&\msf D_{\bc \Omega} \log \mc E_{\mr{tr1}}(\bc \vartheta; \bc\mu, \bc \Omega, \eta, \alpha, \beta) = -\frac{1}{2} (\mr{vec}(\bc \Omega^{-\top}))^\top - \frac{\beta }{2\alpha\eta} (t/\alpha\eta)^{\beta-1}  \msf D_{\bc \Omega} t, \\
& \msf D_{\bc \Omega} t = - (\mr{vec}(\bc \Omega^{-1} (\bc \vartheta - \bc \mu)(\bc \vartheta - \bc \mu)^\top \bc \Omega^{-1})^\top)^\top,
\end{align}
\begin{align}
&\msf D_{\bc \mu} \log \mc E_{\mr{tr1}}(\bc \vartheta; \bc\mu, \bc \Omega, \eta, \alpha, \beta) = -\frac{\beta}{2\alpha\eta} (t/\alpha\eta)^{\beta-1}\msf D_{\bc \mu} t,
\\
& \msf D_{\bc \mu} t= -2 (\mr{vec}((\bc \vartheta - \bc \mu)^\top \bc \Omega^{-1})^\top)^\top,
\end{align}
\begin{align}
& \msf D_{\log \eta} \log \mc E_{\mr{tr1}}(\bc \vartheta; \bc\mu, \bc \Omega, \eta, \alpha, \beta) =  -\frac{d}{2} + \frac{\beta t}{2\alpha\eta} (t/\alpha\eta)^{\beta-1}, 
\end{align}
\begin{align}
& \msf D_{\alpha} \log \mc E_{\mr{tr1}}(\bc \vartheta; \bc\mu, \bc \Omega, \eta, \alpha, \beta) = -\frac{d}{2\alpha} + \frac{\beta t}{2\eta\alpha^2} (t/\alpha\eta)^{\beta-1}, 
\end{align}
\begin{align}
&\msf D_{\beta} \log \mc E_{\mr{tr1}}(\bc \vartheta; \bc\mu, \bc \Omega, \eta, \alpha, \beta) = \msf D_{\beta} \log c(\alpha, \beta) -\frac{1}{2} (t/\alpha\eta)^{\beta} \log(t/\alpha\eta),
\\
& \msf D_{\beta} \log c(\alpha, \beta) = \frac{1}{\beta} + \frac{d}{2\beta^2} ( \psi(d/2\beta) + \log 2). 
\end{align}
\subsubsection{The $\alpha$-derivatives of the stochastic representation}
\label{app:der_mpe_sr}
For the stochastic representation of the mPE distribution given in \eqref{eq:rt_mpe}, we have
\begin{align}
&\msf D_{\bc \mu} \bc \vartheta = \bc I_d,\\
&\msf D_{\bc \Omega} \bc \vartheta = \msf D_{\bc \Phi} \bc \vartheta  \msf D_{\bc \Omega} \bc \Phi = \varsigma \alpha \eta (\bc \nu^\top \otimes \bc I_d) (\bc \Phi^{-1} \otimes \bc I_d),\\
&\msf D_{\alpha} \bc \vartheta = \msf D_{\bc \Phi} \bc \vartheta \msf D_{\alpha} \bc \Phi= \varsigma \eta(\bc \nu^\top \otimes \bc I_d) \mr{vec}(\bc \Omega \bc \Phi^{-\top}),
\\
& \msf D_{\log\eta}\bc \vartheta = \eta \msf D_{\bc \Phi} \bc \vartheta \msf D_{\eta} \bc \Phi= \eta \varsigma \alpha(\bc \nu^\top \otimes \bc I_d) \mr{vec}(\bc \Omega \bc \Phi^{-\top}),
\\
& \msf D_{\beta}\bc \vartheta = \msf D_{\varsigma} \bc \vartheta \msf D_{\beta}\varsigma = -\frac{d}{2\beta^3 \varsigma^{2\beta-1}} (1 + \epsilon ({{2d}/{\beta}})^{-\frac{1}{2}}) \mr{vec}(\bc \Phi \bc \nu).
\end{align}

\section{Derivative of the estimators}
\label{app:estimator}
\subsection{Derivative of the estimator \eqref{eq:est_n}}
The derivatives of the estimator \eqref{eq:est_n} with respect to the generative parameters $\theta$ and the recognition parameters $\phi$ are given by
\begin{align}
& \msf D_\theta \widetilde{\mc L} =  \frac{1}{r}\sum_{l=1}^r \msf D_{\theta}\log~{p_{\theta}(\vc x\h i \given \vc s\h {i,l})}, \\
& \msf D_{\phi} \widetilde{\mc L} =  \Big(\frac{1}{r}\sum_{l=1}^r\msf D_{\vc s\h {i,l}}\log~{p_{\theta}(\vc x\h i \given \vc s\h {i,l})} \msf D_{\phi} \vc s\h {i,l} \Big)  - \msf{D}_{\phi}\Delta_{\mr{KL}}(q_{\phi}(\vc s\mid \vc x\h i) || p_{\pi}(\vc s)),
\end{align}
where $\msf D_{\phi} \vc s\h {i,l}$ can be computed by taking the $\alpha$-derivatives from the stochastic representation of the trace-one Gaussian distribution given by Eq.~\eqref{eq:rep}. These derivatives are computed in Appendix~\ref{app:der_mg_sr}. Other derivatives can be computed by making use of the results in Appendix~\ref{app:der_mg} and Appendix~\ref{app:der_mpe}.

\subsection{Derivative of the estimator \eqref{eq:est_mpe}}
The Derivatives of the estimator \eqref{eq:est_mpe} with respect to the generative parameters $\theta$ and the recognition parameters $\phi$ are computed from
\begin{align}
& \msf D_\theta \widetilde{\mc L} =  \frac{1}{r}\sum_{l=1}^r \msf D_{\theta}\log~{p_{\theta}(\vc x\h i \given \vc s\h {i,l})}, \\
& \msf D_{\phi} \widetilde{\mc L} = \frac{1}{r}\sum_{l=1}^r \msf D_{\vc s\h {i,l}} \left(\log~{p_{\theta}(\vc x\h i \given \vc s\h {i,l})} + \log p_{\pi}(\vc s\h {i,l}) - \log q_{\phi}(\vc s\h {i,l}\mid \vc x\h i)\right) \msf D_{\phi} \vc s\h {i,l}  - \msf D_{\phi}\log q_{\phi}(\vc s\h {i,l}\given \vc x\h i),
\end{align}
where $\msf D_{\phi} \vc s\h {i,l}$ is computed by taking the $\alpha$-derivatives from the stochastic representation of the mPE distribution given by \eqref{eq:rt_mpe}. These derivatives are computed in Appendix~\ref{app:der_mpe_sr}.  Other derivatives can be computed by making use of the results in Appendix~\ref{app:der_mg} and Appendix~\ref{app:der_mpe}.

\section{Additional details on the experiments}
\subsection{Example 1}
\label{app:example1}
\paragraph{Data generation.}
Let $\bc A\in \mbb R^{d_0\times 20} $ be a matrix where each of its elements is generated from a standard normal distribution. The matrix $\bc A$ is kept fixed. 
The $i^{\text{th}}$ class covariance $\vc Y_i$ is computed according to the following procedure:
\begin{enumerate}
\item Draw $10^4$ samples from a known Gaussian distribution $\mc N(\bc \mu_i, \bc \Sigma_i)$ with a unique mean ${\bc \mu_i\in\mbb R^{d_0}}$ and a unique dense covariance matrix ${\bc \Sigma_i\in\mbb R^{d_0\times d_0}}$. 
\item  Let $\bc t_j$ be a random sample from this Gaussian. For this sample, compute $\bc y_j=\bc A \bc t_j$. For all $10^4$ samples, collect the results into $\underline{\bc y}=\{\bc y_j\}_{j=1}^{10^4}$. 
\item Compute the sample covariance of $\underline{\bc y}$ and normalize the resulting covariance matrix to trace one, that is ${\vc Y_i \leftarrow \mr{cov}(\underline{\bc y})/\mr{tr}(\mr{cov}(\underline{\bc y}))}$.
\end{enumerate}
\paragraph{Initialization.}
All models use the same batch size (equal to 5), the same choice of activation matrix function, which is given by the Mercer sigmoid kernel~\eqref{eq:msk}, and the same optimizer (the Adam optimizer \citep{Kingma2014} with default settings).

\subsection{Example~1: Additional analysis}
\label{app:ex1mlp}
The standard MLP was initialized with 3 layers. We varied the number of units and found that it affects the performance only marginally as long as it is sufficiently large. For this analysis we set the number of units to 400. Figure~\ref{fig:example1}-E shows the predicted covariance matrices for the two test samples. The complete results are shown in Table~\ref{tb:ex1_mlp0}.

\begin{table}[h!]
\setlength{\tabcolsep}{8pt}
\renewcommand{\arraystretch}{1.4}
  \tiny
  \caption{SPD matrix learning using the MLP model (refer to Example~1).}
  \begin{center}
  \begin{tabular}{lccc c ccc}
\hline
\multicolumn{4}{c}{$d_0=10, n_{\mr{train}}=20$}{$d_0=20, n_{\mr{train}}=20$} \hspace{-38ex} \\
\cline{2-4} \cline{6-8} 
loss    &\hspace{-3ex}$E_{\mr{quad}}$ &\hspace{-3ex} $E_{\mr{QRE}}$ &\hspace{-3ex} $E_{\mr{Stein}}$ & \hspace{-3ex}~ &\hspace{-3ex} $E_{\mr{quad}}$ &\hspace{-3ex} $E_{\mr{QRE}}$ &\hspace{-3ex} $E_{\mr{Stein}}$\\
\hline
$\ell_{\mr{quad}}$  &\hspace{-3ex} $0.63$  &\hspace{-3ex} $4.3$ &\hspace{-3ex}  $36.7$ &\hspace{-3ex}  &\hspace{-3ex}$0.72$ &\hspace{-3ex}$6.25$ &\hspace{-3ex} $97.4$\\
\hline
\end{tabular} \hspace*{-1.5ex}
  \begin{tabular}{lccc c ccc}
\hline
\multicolumn{4}{c}{$d_0=10, n_{\mr{train}}=100$}{$d_0=20, n_{\mr{train}}=100$} \hspace{-33ex} \\
\cline{2-4} \cline{6-8} 
    &\hspace{-3ex}$E_{\mr{quad}}$ &\hspace{-3ex} $E_{\mr{QRE}}$ &\hspace{-3ex} $E_{\mr{Stein}}$ & \hspace{-3ex}~ &\hspace{-3ex} $E_{\mr{quad}}$ &\hspace{-3ex} $E_{\mr{QRE}}$ &\hspace{-3ex} $E_{\mr{Stein}}$\\
\hline
  &\hspace{-3ex} $0.65$  &\hspace{-3ex} $4.55$ &\hspace{-3ex}  $37.4$ &\hspace{-3ex}  &\hspace{-3ex}$0.73$ &\hspace{-3ex}$6.4$ &\hspace{-3ex} $98.7$\\
\hline
\end{tabular}
\vspace{-0ex}
\end{center}
 \label{tb:ex1_mlp0}
\end{table}

\subsection{Example 2}
\label{app:example2}
\paragraph{Data generation.}
See the data generation procedure in Example~1.
\paragraph{Initialization.}
Both models \eqref{eq:mlps} and \eqref{eq:mlpa} use the same batch size (equal to 5), the same choice of loss function \eqref{eq:loss_qre}, and the same optimizer (the Adam optimizer \citep{Kingma2014} with default settings). Both models use the same choice of the output activation matrix function, given by the Mercer sigmoid kernel \eqref{eq:msk}. The model in \eqref{eq:mlps} uses the hyperbolic tangent as the activation function across the hidden layers, while~\eqref{eq:mlpa} makes use of the same choice of the activation matrix function as in its output layer.

\subsection{Frey-Face experiment}
For all models listed in Table~\ref{tb:abb}, the same initialization is used. We use a batch size of $10$, and as before, we use the Adam optimizer with its default settings. 
\section{Numerical evaluation of the quality of approximation used in the stochastic representation of the mPE distribution.}
\label{app:gamma_vs_gauss}
We generated random samples from an mPE distribution with known parameters. For generating the random samples, we used the stochastic representation of the distribution, once through 
\begin{align}
	{\varsigma^{2\beta}\sim \mc G(\frac{d}{2\beta}, 2)}, 
\end{align}
as in \eqref{eq:gamma_app} and the other time through 
\begin{align}
	{\varsigma^{2\beta}\sim\mc N(\frac{d}{\beta}, \frac{2d}{\beta})},
\end{align}
as in \eqref{eq:gauss_app}. Since the true parameters of the distribution are known, we can compute analytically the exact mean vector and the exact covariance matrix using \eqref{eq:meanandcov}. We can also compute the sample mean vector and the sample covariance matrix from the generated random samples. This would allow us to compare them against the exact mean and the exact covariance of the distribution from which samples were generated.

\begin{figure}[t!]
\vspace*{-1ex}
  \begin{center}
    \includegraphics[width=0.99\textwidth]{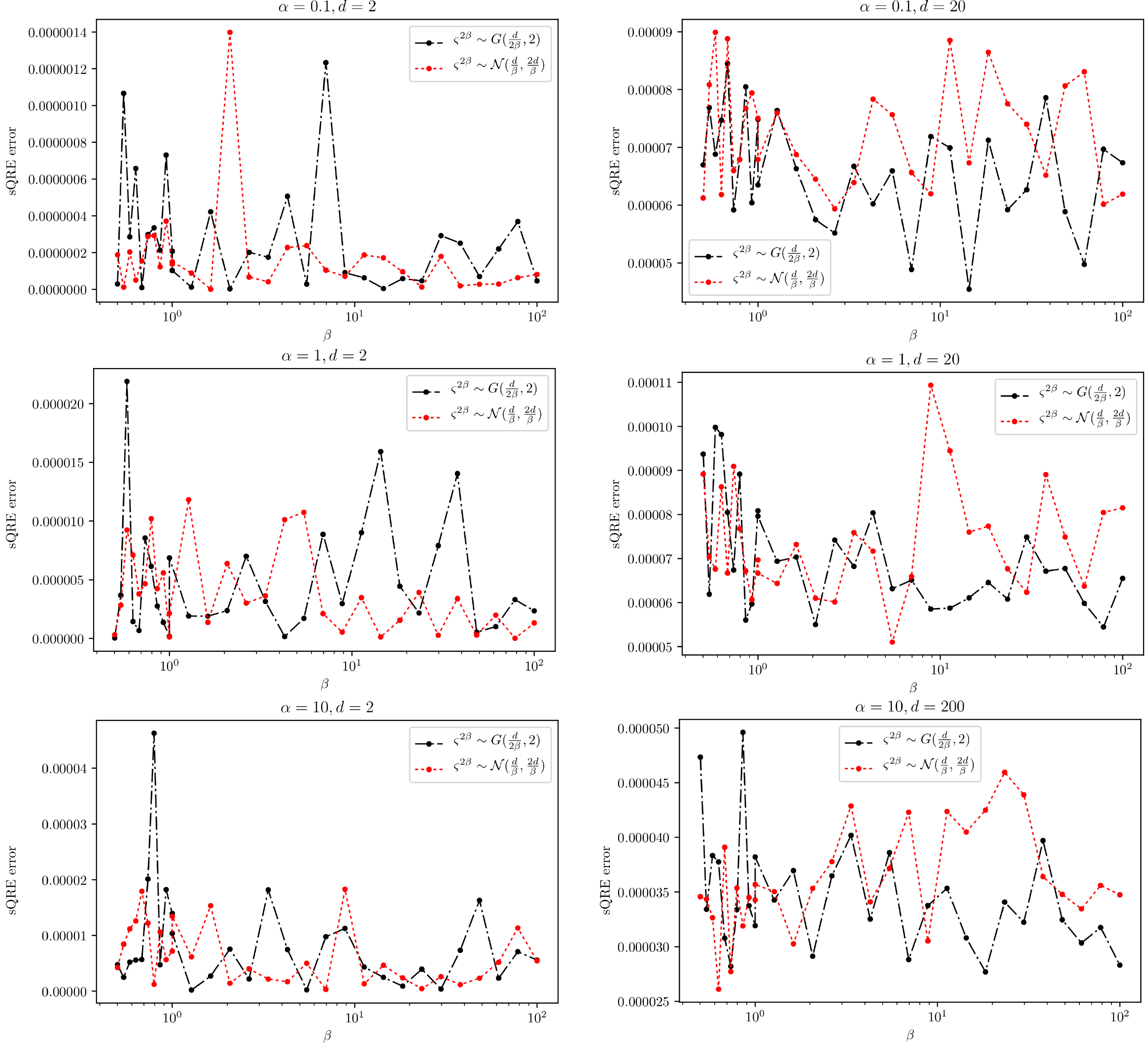}
  \end{center}
  \caption{The sQRE error (the symmetrized von Neumann divergence given by \eqref{eq:loss_qre}) between the true covariance matrix and the sample covariance matrices computed from the stochastic representation of the mPE distribution via ${\varsigma^{2\beta}\sim \mc G(\frac{d}{2\beta}, 2)}$ and ${\varsigma^{2\beta}\sim\mc N(\frac{d}{\beta}, \frac{2d}{\beta})}$. Sample covariances are computed from $10^5$ samples.}
  \vspace*{-2ex}
  \label{fig:gamma_vs_gauss_cov}
  \end{figure}

\begin{figure}[t!]
\vspace*{-1ex}
  \begin{center}
    \includegraphics[width=0.99\textwidth]{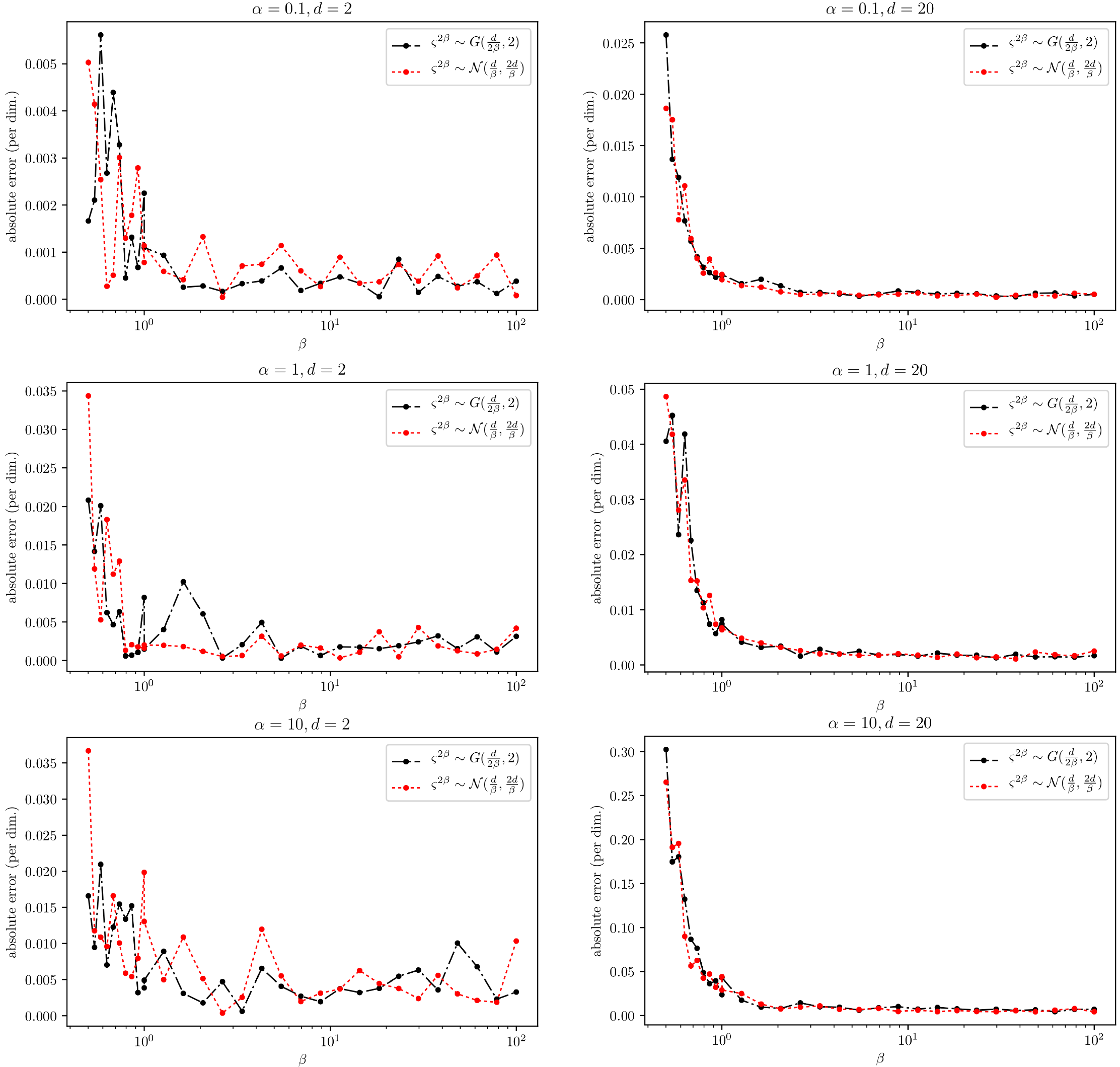}
  \end{center}
  \caption{The absolute error per dimension between the true mean vector and the sample mean vector computed from the stochastic representation of the mPE distribution via ${\varsigma^{2\beta}\sim \mc G(\frac{d}{2\beta}, 2)}$ and ${\varsigma^{2\beta}\sim\mc N(\frac{d}{\beta}, \frac{2d}{\beta})}$. Sample mean vectors are computed from $10^5$ samples.}
  \vspace*{-2ex}
      \label{fig:gamma_vs_gauss_mean}
  \end{figure}

For comparing two covariance matrices, we use the symmetrized von Neumann divergence, also commonly known as the symmetrized quantum relative entropy (sQRE) given by \eqref{eq:loss_qre}. For comparing the mean vectors, we simply use the absolute error per dimension. The experiment is carried out for various $\alpha$, $\beta$, and $d$. Figures~\ref{fig:gamma_vs_gauss_cov} and \ref{fig:gamma_vs_gauss_mean} summarize the results. The Gaussian approximation in \eqref{eq:gauss_app} performs quite well in general. 
\clearpage
\section{Supplementary figures}
\label{app:figs}
  \begin{figure}[h]
\vspace*{-1ex}
  \begin{center}
    \includegraphics[width=.99\textwidth]{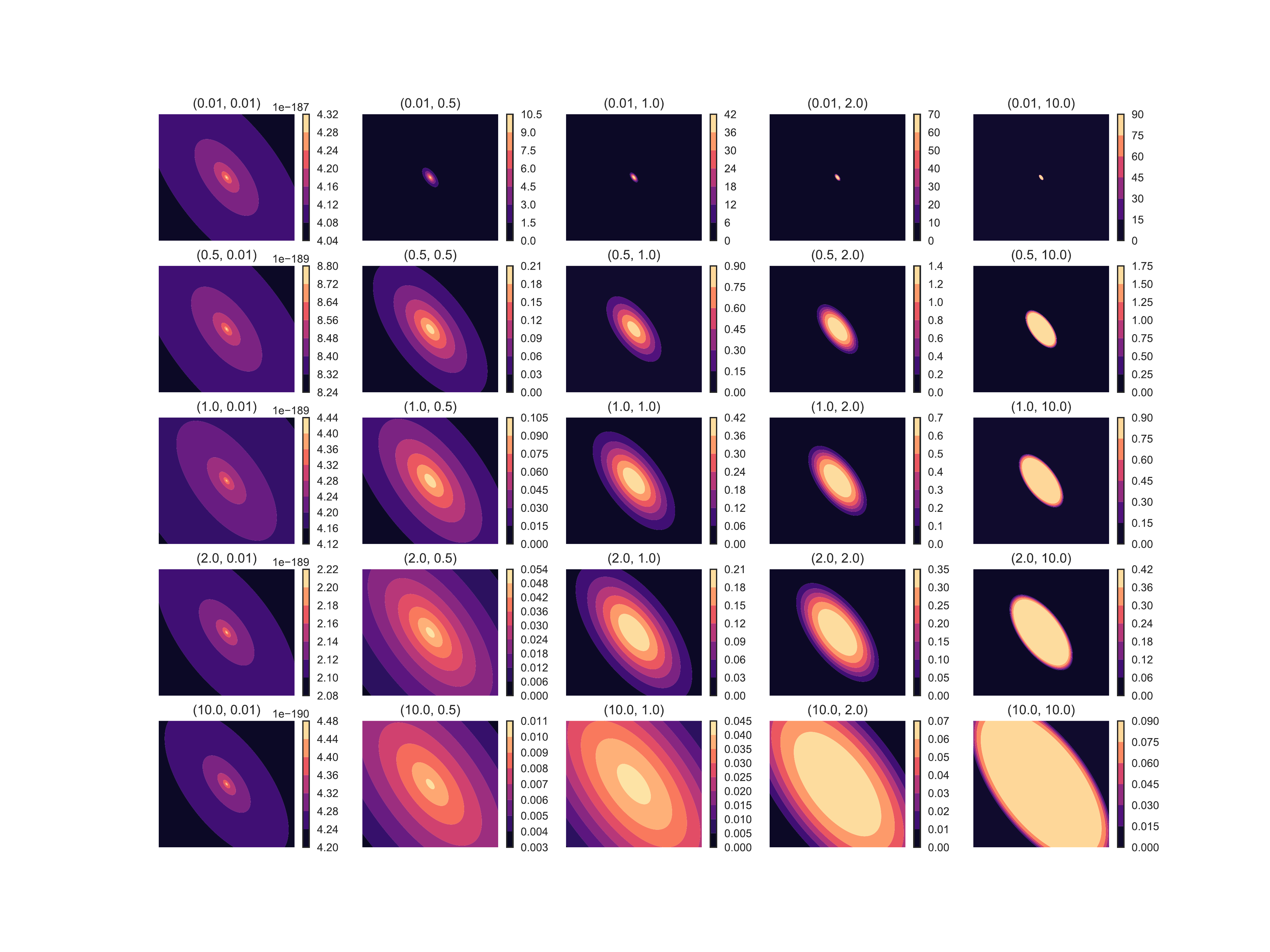}
  \end{center}
  \vspace*{-3ex}
  \caption{The probability density function of a trace-one mPE distribution, ${\mc E_{\mr{tr1}}(\bc \mu, {\bc M}/{\eta}, \eta, \alpha, \beta)}$ for fixed values of ${\bc \mu, \bc M, \eta=\mr{tr}(\bc M)}$ and varying values of scale and shape parameters $(\alpha, \beta)$. When ${\alpha=1}$ and $\beta=1$, the density corresponds to the multivariate Gaussian distribution ${\mc N(\bc \mu, \bc M)}$.}
  \label{fig:pdf}
  \end{figure}
  \begin{figure}[h!]
\vspace*{-1ex}
  \begin{center}
    \includegraphics[width=0.60\textwidth]{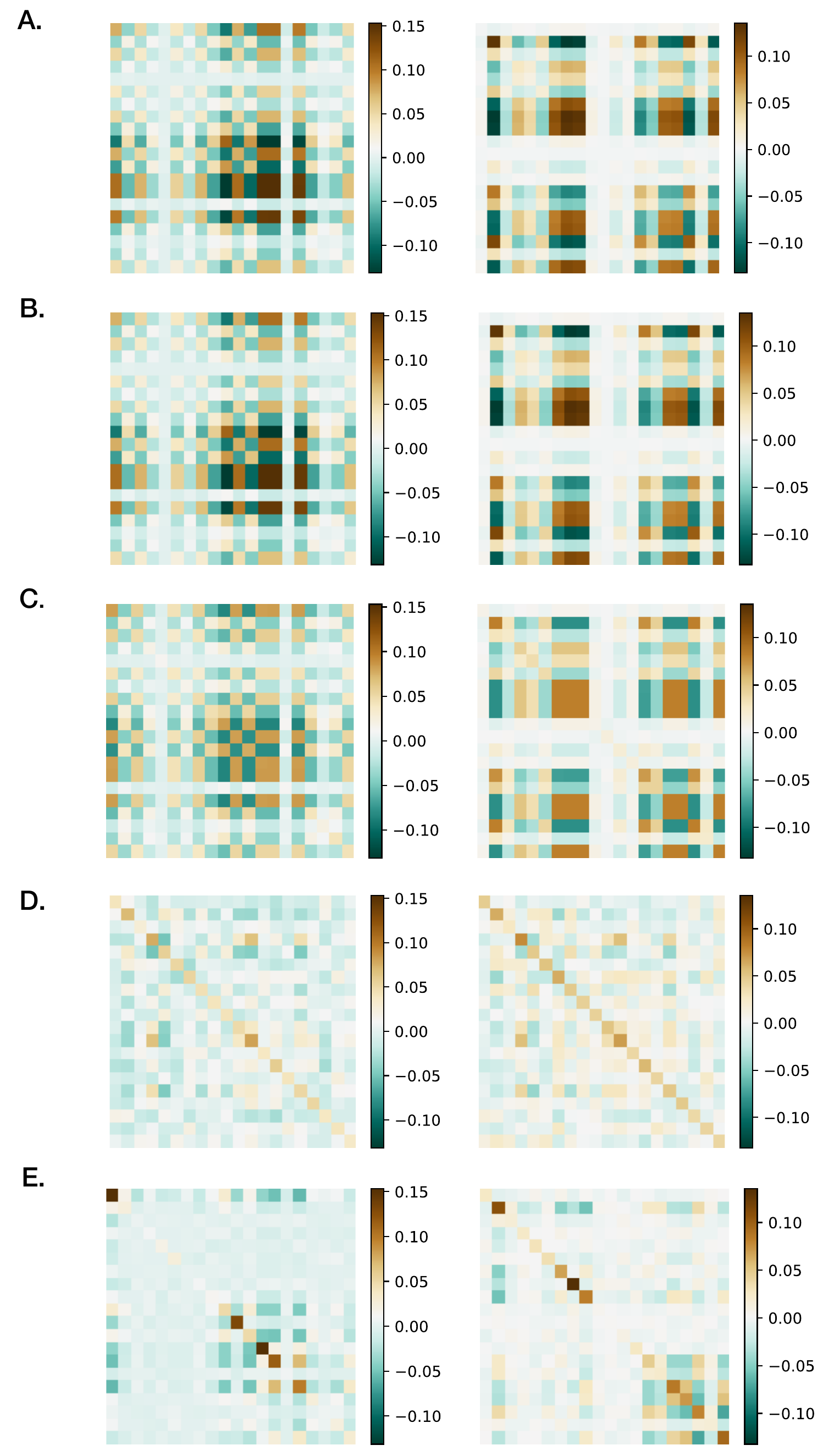}
  \end{center}
  \caption{SPD matrix learning using the mMLP (see Example~1 in Section~\ref{sec:example1}). (A) Two instances of target covariance (SPD) matrices $(20\times 20)$. (B) Estimated covariance matrices by the mMLP using $\ell_{\mr{QRE}}$, (C) using $\ell_{\mr{quad}}$, (D) using $\ell_{\mr{Stein}}$. (E) Estimated covariance matrices using the vanilla MLP and quadratic loss.}
  \vspace*{-2ex}
    \label{fig:example1}
  \end{figure}

  \begin{figure}[t!]
\vspace*{-1ex}
  \begin{center}
    \includegraphics[width=0.90\textwidth]{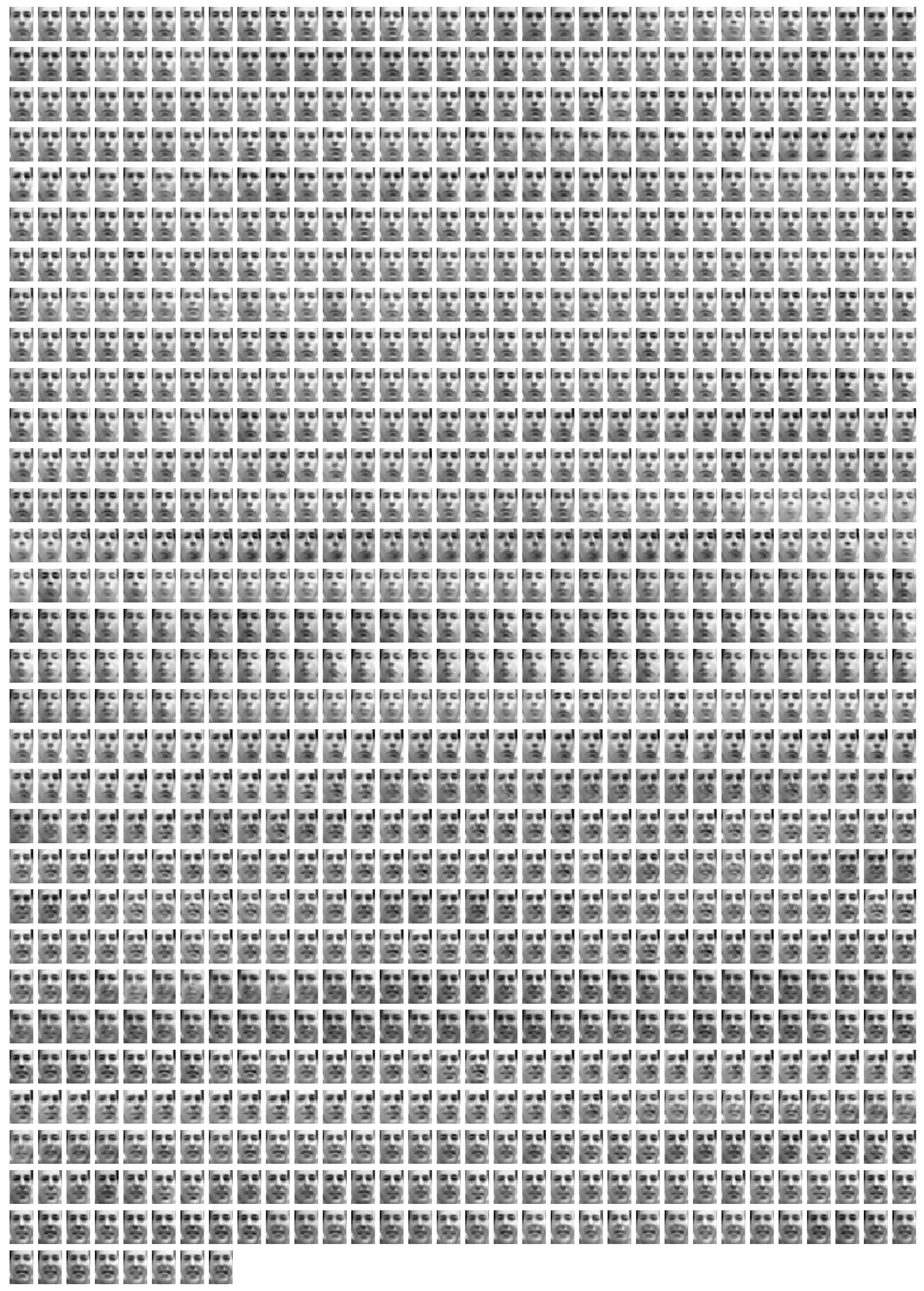}
  \end{center}
  \caption{Frey Face training set (1000 samples). Note that these images are in fact reconstructed from the first $10$ principal components.}
  \vspace*{-2ex}
    \label{fig:train}
  \end{figure}

  \begin{figure}[t!]
\vspace*{-1ex}
  \begin{center}
    \includegraphics[width=0.90\textwidth]{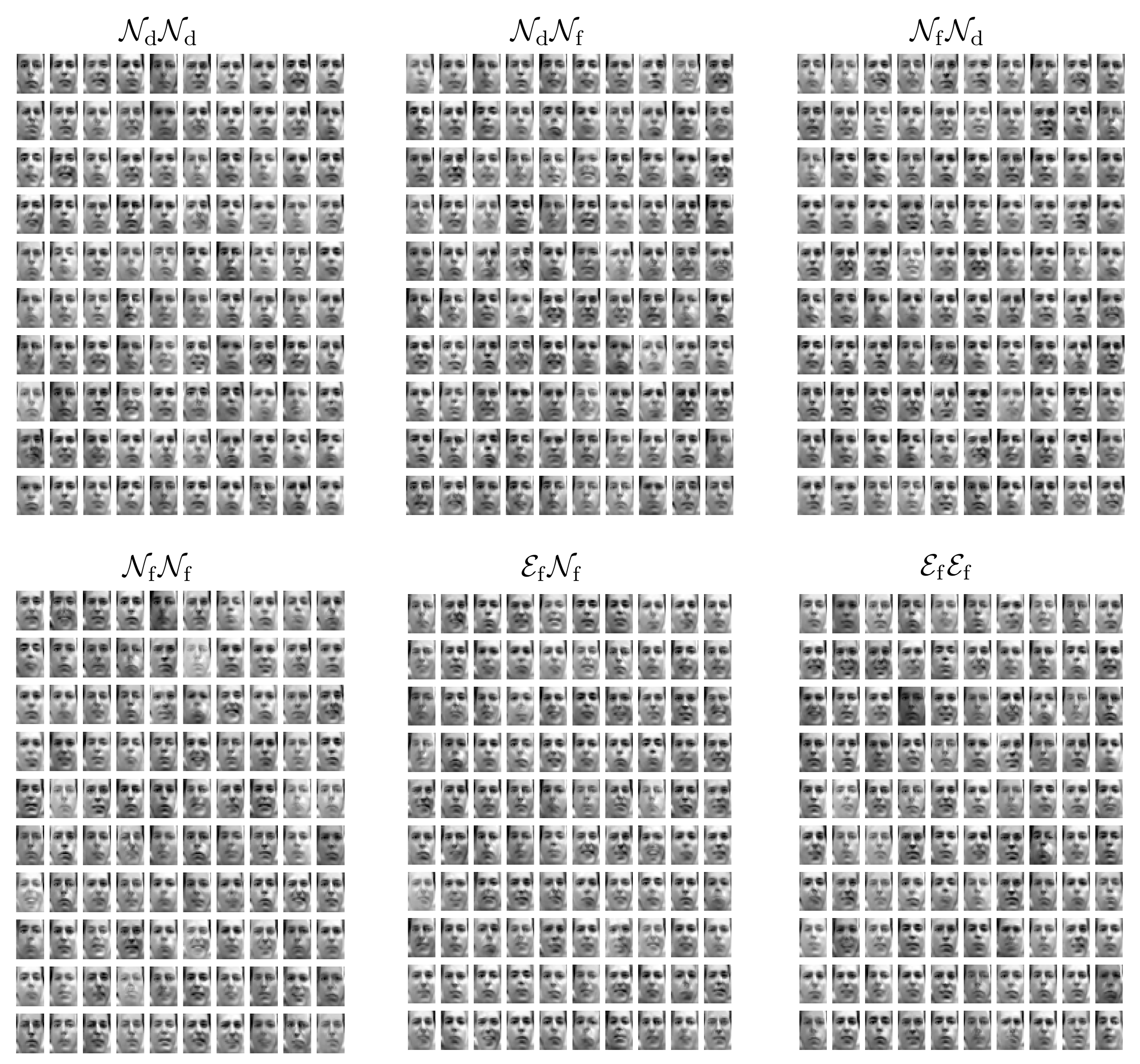}
  \end{center}
  \caption{Random samples generated from the models listed in Table~\ref{tb:abb} using $5$ latent variables (100 random samples for each model). The models are trained on 1000 samples from the training set shown in Figure~\ref{fig:train}.}
  \vspace*{-2ex}
    \label{fig:5}
  \end{figure}

  \begin{figure}[t!]
\vspace*{-1ex}
  \begin{center}
    \includegraphics[width=0.90\textwidth]{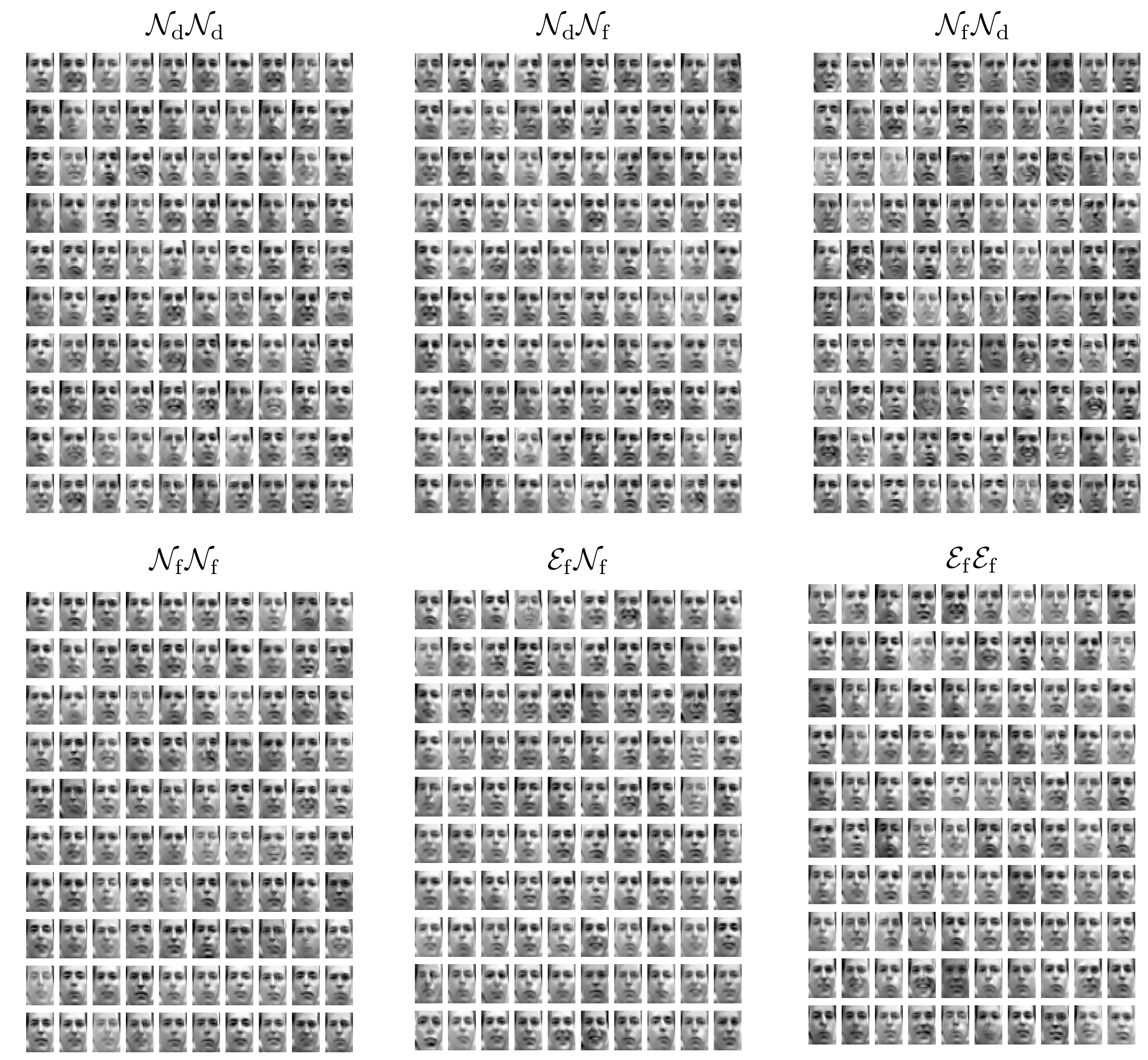}
  \end{center}
  \caption{Random samples generated from the models listed in Table~\ref{tb:abb} using $8$ latent variables (100 random samples for each model). The models are trained on 1000 samples from the training set shown in Figure~\ref{fig:train}.}
  \vspace*{-2ex}
    \label{fig:8}
  \end{figure}
\end{document}